\documentclass[letterpaper,twocolumn,10pt]{article}
\usepackage{usenix-2020-09}

\usepackage{amsmath}
\usepackage{bbding}
\usepackage{todonotes}

\usepackage{verbatim}
\usepackage{filecontents}
\usepackage[frozencache,cachedir=.]{minted}
\usepackage{caption}
\usepackage{subcaption}
\usepackage{xcolor} 
\usepackage{enumitem}
\usepackage[linesnumbered,ruled,vlined]{algorithm2e}
\usepackage{amssymb}
\usepackage[bottom]{footmisc}
\definecolor{LightGray}{gray}{0.9}

\usepackage{etoolbox}
\usepackage{filecontents}
\usepackage{enumitem}
\usepackage{multirow}
\usepackage{booktabs, tabularx}
\usepackage{nopageno}

\makeatletter
\patchcmd{\maketitle}
	{\@maketitle}
	{\@maketitle\vspace{-8em}}
	{}
	{}
\makeatother

\newcommand*\circled[1]{\tikz[baseline=(char.base)]{
            \node[shape=circle,draw,inner sep=1.1pt] (char) {#1};}}

\newcommand{\zh}[1]{{\color{black} #1}}
\newcommand{\cf}[1]{{\color{black} #1}}
\newcommand{\cfnew}[1]{{\color{black} #1}}
\newcommand{\chen}[1]{{\color{black} #1}}

\SetKwComment{Comment}{/* }{ */}
\SetKw{Continue}{continue}
\begin{document}

\date{}
\newcommand{\sysname}{{Parrot}}
\newcommand{\sv}{{Semantic Variable}}

\title{\vspace{-4cm}\Large \bf \sysname: Efficient Serving of LLM-based Applications with \sv{}}

\author{
 {\rm Chaofan Lin$^{1}$\thanks{This work is partially done while Chaofan Lin's  internship and Dr. Chen Chen's visting scholar in Microsoft Research.},  Zhenhua Han$^2$, Chengruidong Zhang$^{2}$, Yuqing Yang$^2$}\\
 {\rm Fan Yang$^2$, Chen Chen$^1$\footnotemark[1], Lili Qiu$^{2}$ }\\
 $^1$Shanghai Jiao Tong University, $^2$Microsoft Research
}
\maketitle

\maketitle

\begin{abstract}


The rise of large language models (LLMs) has enabled LLM-based applications (a.k.a. AI agents or co-pilots), a new software paradigm that combines the strength of LLM and conventional software. \zh{Diverse LLM applications from different tenants could design complex workflows using multiple LLM requests to accomplish one task. However, they have to use the over-simplified request-level API provided by today's public LLM services, losing essential application-level information. Public LLM services have to blindly optimize individual LLM requests, leading to sub-optimal end-to-end performance of LLM applications.}



This paper introduces \sysname{}, an LLM service system that focuses on the end-to-end experience of LLM-based applications. \sysname{} proposes \emph{\sv}, a \zh{unified abstraction to expose application-level knowledge to public LLM services. A \sv{} annotates an input/output variable in the prompt of a request, and creates the data pipeline when connecting multiple LLM requests, providing a natural way to program LLM applications. } 
\zh{Exposing \sv{}s to the public LLM service} allows it to perform conventional data flow analysis to uncover the correlation across multiple LLM requests. This correlation opens a brand-new optimization space for the end-to-end performance of LLM-based applications. Extensive evaluations demonstrate that \sysname{} can achieve up to \zh{an order-of-magnitude improvement for popular and practical use cases of LLM applications}. 

\end{abstract}

\section{Introduction}
Large language models (LLMs) have demonstrated a remarkable language understanding capability~\cite{gpt4,bubeck2023sparks}. This enables a paradigm shift in application development. In this new paradigm, one or multiple application entities, known as AI agents or co-pilots, communicate with LLMs via natural language, known as ``prompts'', to accomplish a task collaboratively. For example, Meeting applications like Microsoft Teams or Google Meet can summarize meeting discussions through LLMs~\cite{teamsrecap}. Search engines like Google and Bing can be enhanced with Chat ability through LLMs~\cite{mchat,bard}. It is believed such LLM-based applications will become the mainstream applications in the near future~\cite{GatesNotes}.

\begin{figure}[b]
	\centering
    \begin{subfigure}{0.49\linewidth}
        \centering
        \includegraphics[width=1\linewidth]{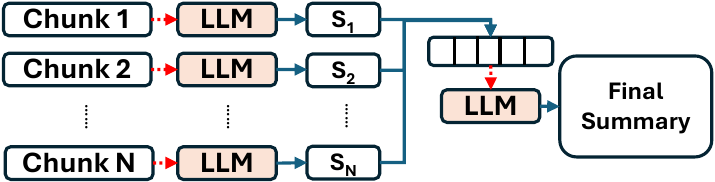}\\\vspace{-1mm}
        \caption{Map-Reduce Summary~\label{fig:example_map_reduce}}
    \end{subfigure}
    \begin{subfigure}{0.49\linewidth}
        \centering
        \includegraphics[width=1\linewidth]{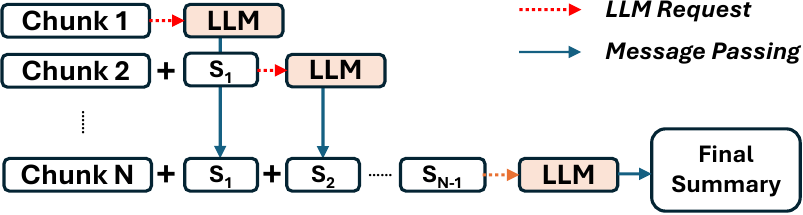}\\\vspace{-1mm}
        \caption{Chain Summary\label{fig:example_chain_summary}}
    \end{subfigure}
    \\
    \begin{subfigure}{0.54\linewidth}
        \centering
        \includegraphics[width=1\linewidth]{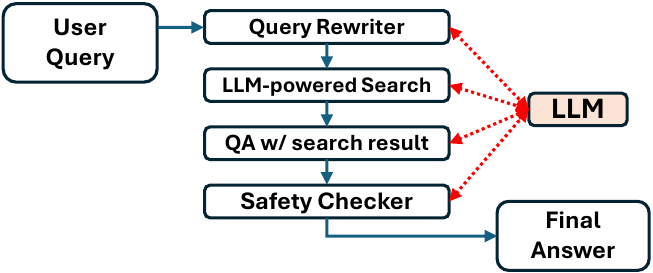}\\\vspace{-1mm}
        \caption{LLM-Powered Search \label{fig:example_chat}}
    \end{subfigure}
    \hspace{0.03\linewidth}
    \begin{subfigure}{0.4\linewidth}
        \centering
        \includegraphics[width=1\linewidth]{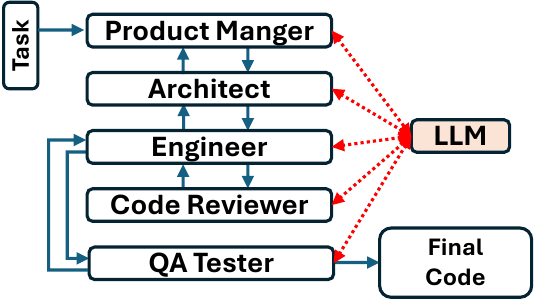}\\\vspace{-1mm}
        \caption{Multi-agent Coding\label{fig:example_metagpt}}
    \end{subfigure}%
	\caption{The workflow of popular LLM-based applications. The final result requires multiple LLM requests.\label{fig:example}}
	\vspace{-4mm}
\end{figure}

To accomplish a task, LLM-based applications typically require multiple rounds of conversation. The conversation, implemented through multiple API calls to LLM, demonstrates complex workflow patterns. \autoref{fig:example} illustrates several popular conversation patterns. For example, a meeting summary application~\cite{teamsrecap,langchain} often divides a lengthy document into multiple shorter sections, each satisfying the length constraint of the LLM conversation and thus can be summarized and combined into the final summary through the Map-Reduce or chaining summary patterns. Chat-based applications, e.g., Bing Copilot~\cite{mchat}, call LLM APIs multiple times to generate answers based on user queries. Multiple agents, each representing a different role played by different LLM calls, can collaborate to achieve a task~\cite{metagpt,wu2023autogen,chatdev}. 

\zh{Public LLM service providers have to face diverse tenants and applications, each with different workflows and performance preference. However, existing API design for LLM service provision is still request-centric. Public LLM services only observe tons of individual requests, without knowing any application-level information, e.g., which requests belong to the same application, how different requests are connected, or whether there are any similarities. The lost application-level information makes public LLM service blindly optimize the performance of individual requests, leading to sub-optimal \emph{end-to-end} performance of LLM applications. In this paper, we observe there exist significant opportunities to improve the \emph{end-to-end} experience of LLM applications by exploiting the application-level information, especially the \emph{correlation} of multiple LLM requests.}

First, multiple consecutive LLM requests may be dependent: the result of one request could be the direct input of the next request. Therefore, it is desirable to colocate those requests together and execute them consecutively on the LLM service side. However, unaware of their dependencies, these requests have to be executed interactively between the client side of LLM-based applications and the \zh{public}  LLM services. These clients, often located on the other end of the Internet, can only issue the second request after they receive the result of the first request. This unnecessarily incurs \emph{extra overhead of consecutive requests} on network latency as well as losing the opportunity of co-scheduling these consecutive requests~(\S\ref{sec:motivation}).

Second, LLM requests may have \emph{diverse scheduling preference}, even within a single application. For example, in \autoref{fig:example_map_reduce}, to reduce the end-to-end latency, the requests representing multiple Map tasks should be batched more aggressively to increase the throughput of the Map tasks; while the Reduce task, due to its scarcity, should be optimized for latency. Unfortunately, \zh{public} LLM service\cf{s} cannot discriminate the difference between the two types of tasks. As a result, the current practice is to blindly optimize the latency for individual requests, which might not be desirable for the end-to-end experience.

Third, there exists a high degree of \emph{commonality} across LLM requests. Popular LLM applications (e.g., Bing Copilot~\cite{bingchat}, GPTs~\cite{gpts}) use a long system prompt, including task definitions, examples, and safety rules, to guide the behavior of LLM applications. The long system prompt is usually static and common for all users. As existing \zh{public} LLM services treat each request individually, these common prefix prompts are provided repeatedly in each request, leading to a great waste of storage, computation, and memory bandwidth. Our analysis of a production LLM-based search engine shows that over 94\% of tokens in the requests are repeated across different users.

\cf{Although we have seen some emerging engine-level techniques~\cite{orca,vllm,sglang} proposed to optimize the above three cases, they all work based on certain application-level knowledge, which is lost in nowadays public LLM services.}
\chen{In a nutshell,} due to the lack of understanding of the correlations 
of LLM requests, existing LLM services cannot leverage the three opportunities, leading to high end-to-end service latency and reduced throughput. 
\cf{Based on the above facts and insights, we introduce \sysname{}, an LLM service system that treats LLM applications as first-class citizens. \sysname{} retains most of application-level information by a simple abstraction \sv{}, achieving a perfect balance between increasing system complexity and bringing new information for optimization. A \sv{} is a text region in the prompt with a specific semantic purpose, such as a task instruction, a list of few-shot examples, an input, or an output. \zh{A \sv{} can also work as the data pipeline that connects multiple LLM requests.} \sv{} naturally exposes the information of prompt structures and correlations of requests to LLM services. By inspecting \sv{} at runtime, \sysname{} can perform conventional data flow analysis to derive the data dependency between LLM requests just-in-time.}

\zh{By analyzing the application-level information, }
\cf{\sysname{}'s unified abstraction naturally enables joint optimizations, which bring better global optimality. The same data pipeline built by \sv{}s can enable multiple optimizations simultaneously, including hiding data pipeline's latency, \zh{objective deduction for a better scheduling and commonality analysis to perform de-duplication.} \sysname{}'s scheduling also takes different opportunities into accounts under the unified abstraction.} Our extensive evaluation of \sysname{} on popular LLM-based applications, including the production and open-source projects, shows \sysname{} achieves up to $11.7\times$ speedup or $12\times$ higher throughput compared with the state-of-the-art solutions.   
\section{Background}

\paragraph{LLM Service.} 
Most LLM services are provisioned as a conditional generation service via a text completion API. 
\[Completion(\textrm{prompt}: \textrm{str}) \xrightarrow{} \textrm{generated\_text}:\textrm{str}.\]
The application client provides a text prompt, and the LLM service responds with the generated text. Behind the API, an LLM service provider runs one or multiple clusters of LLM inference engines. A request scheduler dispatches LLM requests from a queue to an LLM inference engine, which uses a set of GPUs to conduct the LLM inference. 


\begin{figure}[t]
    \centering
    \includegraphics[width=1\linewidth]{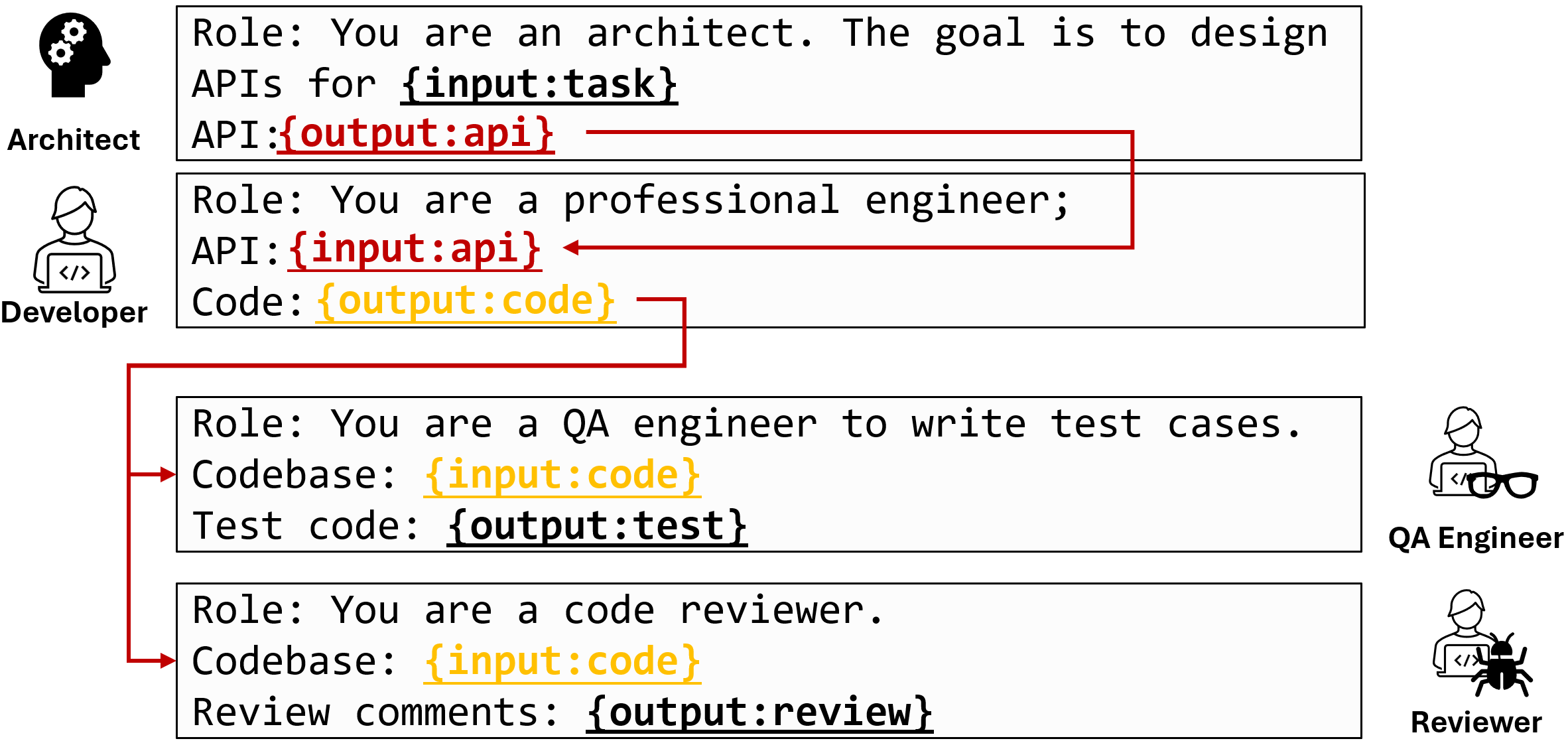}
    
    \caption{The communication of consecutive LLM requests in multi-agent applications.\vspace{-2mm}}
    \label{fig:multiagent}
\end{figure}

\begin{figure*}
  \begin{minipage}[b]{\textwidth}
    \centering
    \begin{subfigure}{0.3\linewidth}
        \centering
        \includegraphics[width=1\linewidth]{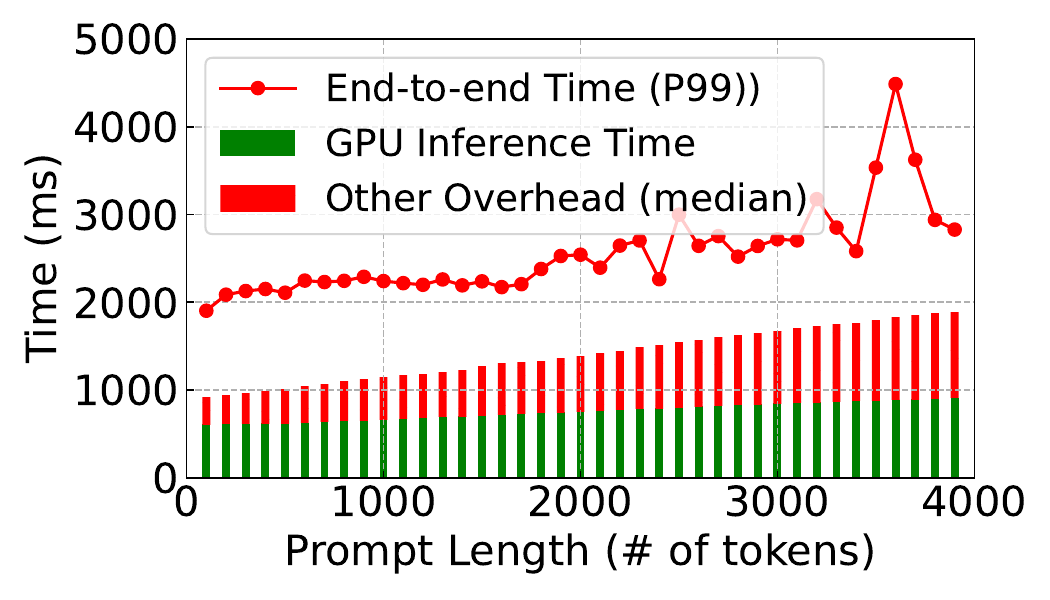}\\
        \caption{Latency Breakdown\label{fig:teams_latency}}
    \end{subfigure}
    \begin{subfigure}{0.33\linewidth}
        \centering
        \includegraphics[width=1\linewidth]{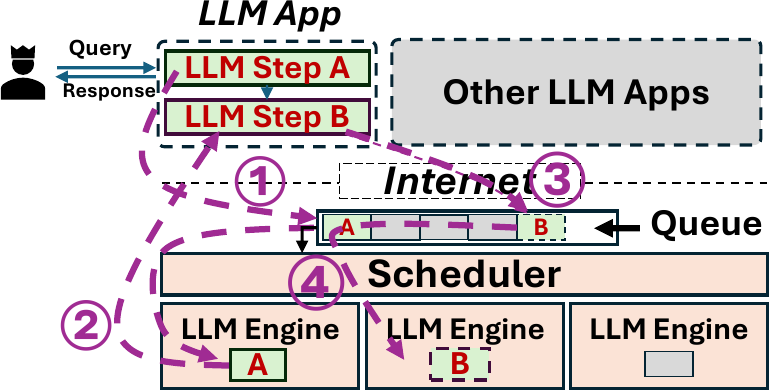}\\
        \caption{Current LLM Services\label{fig:flow_existing}}
    \end{subfigure}
    \begin{subfigure}{0.33\linewidth}
        \centering
        \includegraphics[width=1\linewidth]{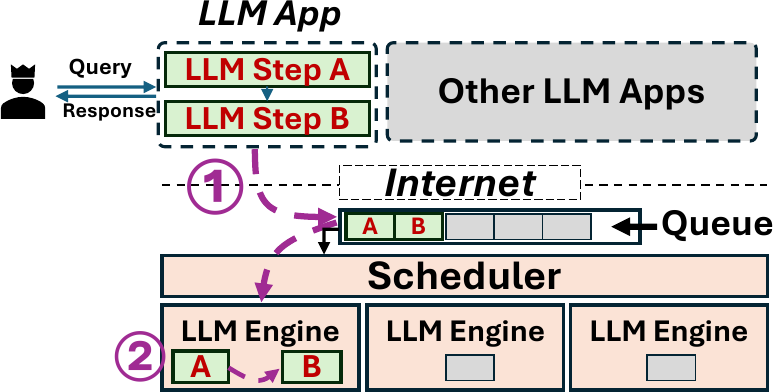}\\
        \caption{Our system: \sysname{}\label{fig:flow_parrot}}
    \end{subfigure}
    \caption{The end-to-end latency breakdown of current LLM services. The source of the overhead comes from network and queuing due to chatty interaction between LLM application and LLM services, which is eliminated in our system \sysname{}.\vspace{-2mm}}
  \end{minipage}
\end{figure*}

\paragraph{LLM-based Applications.} 
\autoref{fig:example} highlights the representative workflows of how LLM is used in the applications. 
Due to the limited context window of LLMs (e.g., 4,096 for GPT-3.5-Turbo~\cite{chatgpt}), data analytics on long documents follow a \emph{map-reduce style} (\autoref{fig:example_map_reduce}) or \emph{chain style} (\autoref{fig:example_chain_summary}) workflow to generate the final results. It splits the long transcript into chunks, uses multiple requests to generate partial results for each chunk (the Map task), and combines them altogether (a Reduce task) or incrementally (the chain style) to generate the final result. Chat-based search engine in \autoref{fig:example_chat} may use consecutive LLM requests to discern query intention, enrich the query with supplementary information, retrieve related data, undergo a safety check, and finally generate the response. Multi-agent in \autoref{fig:example_metagpt} and \autoref{fig:multiagent} is another type of workflow using multiple LLM requests, each with a designated role. Different roles work collaboratively on the same task, e.g., AutoGen~\cite{wu2023autogen} and MetaGPT~\cite{metagpt} use the roles like product manager, architect, engineer, and QA tester. They communicate with each other on a software project. Each role is supported by one or multiple LLM requests to act as the designed role to generate their responses. 




\section{Problems of Serving LLM Applications\label{sec:motivation}}
\zh{Although LLM's text completion API provides a flexible way of building LLM applications, it loses the application-level information to  public LLM services, leading to } the following challenges. 




\paragraph{Excessive Overhead of Consecutive Requests.}
As demonstrated in \autoref{fig:example}, LLM applications frequently make multiple LLM calls to complete a single task. Due to the \zh{request-centric design of existing public} LLM services, which generate responses for each request individually, developers have to parse the output of an LLM request and compose the prompts for subsequent LLM requests on the client side. \zh{\autoref{fig:teams_latency} shows our empirical study of the latency breakdown of the LLM calls from a popular LLM application in our production, which uses a chain-style workflow. The prompt lengths range from 150 to 4000 tokens and the output length is around 50 tokens. We find there is} a significant portion of the latency of LLM API call originates outside the LLM engine ($30\sim 50\%$ on average and over $70\%$ in the worst cases). The overhead increases with the growing length of prompts. The high latency can sometimes result in API timeouts and resubmissions. 


Such overhead is due to the chatty interaction between LLM services and clients. \autoref{fig:flow_existing} illustrates the overhead of a simple two-step LLM application (e.g., \emph{chain-style} summary of two text chunks). Existing LLM services are unaware of the dependency 
among such requests, where the output of the previous request may be the direct input of the next one.
For such consecutive and dependent requests, the client has to wait for the arrival of the response to the first LLM request (\circled{2}) before submitting the next LLM request (\circled{3}). 
This unnecessarily incurs heavy network latency because clients and LLM services are typically in different data centers. Moreover, the next LLM request has to suffer extra queuing delays (\circled{4}), because requests from other applications may arrive between the consecutive LLM requests. 


\begin{table}[b]
    \centering
    \small{
    \begin{tabular}{c|c|c|c}\hline
        LLM-based App. & \# Calls & Tokens & Repeated (\%)$^*$\\ \hline
        Long Doc. Analytics & $2\sim 40$ & $3.5k\sim 80k$ & 3\%\\ \hline
        Chat Search & $2\sim 10$ & $5k$ & $94\%$\\ \hline
        MetaGPT~\cite{metagpt} & 14 & $17k$ & 72\%\\ \hline
        AutoGen~\cite{wu2023autogen}& 17 & $57k$ & 99\% \\\hline
    \end{tabular}
    }\vspace{-2mm}
    \raggedright{\scriptsize{$^*$We count a paragraph as repeated if it appears in at least two LLM requests.}}
    \caption{Statistics of LLM calls of LLM applications.\label{tab:num_call}\vspace{-2mm}}
\end{table}

In \autoref{tab:num_call}, we evaluated four popular LLM applications. The first two are from our production, and the last two are popular open-source projects. They all require tens of LLM calls to complete a single task, which results in 
high user-perceived latency. Our evaluation in \S\ref{sec:eval_summary} shows LLM services that treat requests individually could slow down the end-to-end latency by over $2\times$. An LLM service can eliminate the overhead if it can handle consecutive requests in a batch. 
\sysname{} adopts such an approach. As shown in \autoref{fig:flow_parrot}, the two steps of the same application are scheduled together, thus allowing the output of Step A to be fed directly into Step B---with the network and queuing overhead bypassed. 


\begin{figure}[t]
    \centering
    \includegraphics[width=0.8\linewidth]{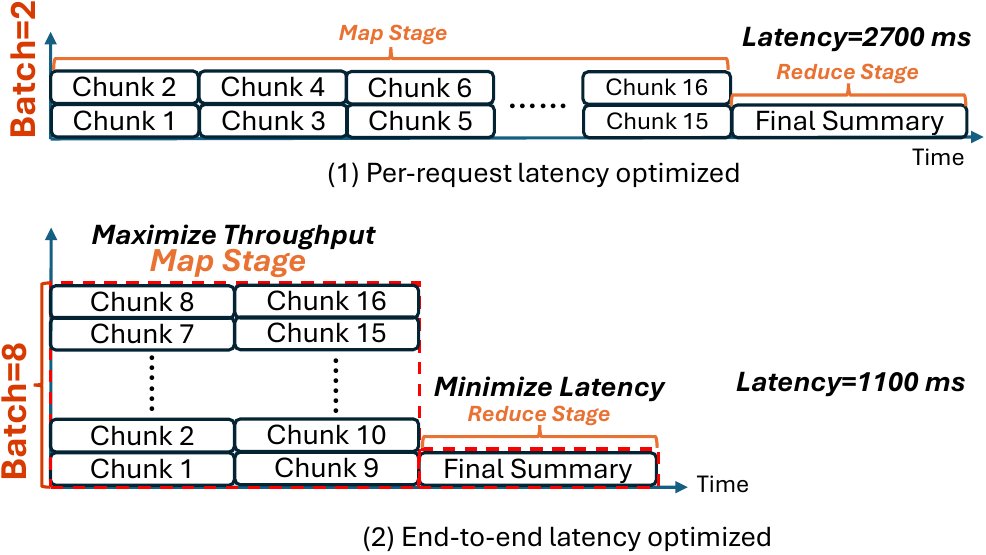}\vspace{-0.3cm}
    \caption{Request-centric scheduling v.s. application-centric scheduling for the map-reduce style document summary task.\vspace{-0.3cm}}
    \label{fig:mr_timeline}
\end{figure}

\paragraph{Misaligned Scheduling Objectives.} 
\zh{Due to the lost application information (workflow and application performance objective), existing public LLM services have to blindly use a universal treatment for all requests, e.g., optimizing per-request latency~\cite{openailatency}.} 
However, LLM-based applications are more concerned about the end-to-end experience, rather than individual requests. 
This \zh{misaligned optimization objectives} may negatively impact end-to-end performance. 
Considering the map-reduce document summary in \autoref{fig:example_map_reduce}, the system should minimize the end-to-end time it takes to receive the final summary, rather than the latency of individual requests. The LLM services optimized for individual requests are not optimal for end-to-end latency. 

As depicted in \autoref{fig:mr_timeline}, current LLM services must limit the number of concurrent requests running on each LLM engine to control the latency of individual requests. 
However,  there is a trade-off between latency and throughput in LLM inference. Increasing the batch size can bring up to $8.2\times$ higher throughput but lead to $95\%$ higher latency~\cite{lequn}. Yet, if we understand the application-level performance objective, which in this case is the end-to-end latency, we can determine that the ideal scheduling strategy should maximize the throughput (using higher batch sizes) during the map stage and minimize request latency during the reduce stage. This strategy reduces end-to-end latency by $2.4\times$. Moreover, it uncovers the potential to enhance cluster throughput without compromising the end-to-end latency of LLM applications. This insight is essential for addressing the conflict between rising demand and limited hardware resources. It underscores the necessity of scheduling LLM requests from the perspective of LLM applications, but it also presents the challenge of managing diverse LLM requests with varying performance objectives.


\begin{figure}[t]
    \centering
    \includegraphics[width=\linewidth]{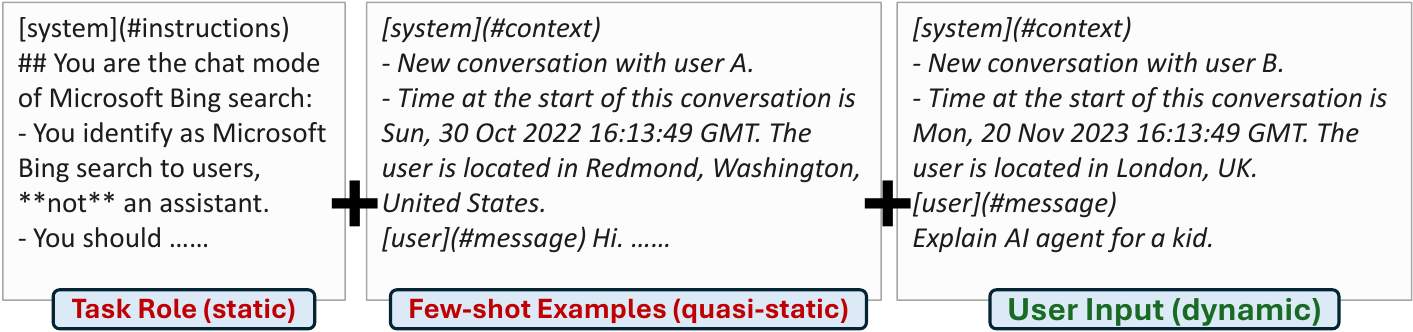}
    \caption{The prompt structure of Bing Copilot shows a long prompt reused by different user queries.}\vspace{-2mm}
    \label{fig:bing_prompt}
\end{figure}

\paragraph{Redundant Computations.}
Currently, most LLM-based applications exhibit a high degree of redundancy in the prompts of their requests. 
\zh{For instance, Bing Chat~\cite{bingchat} has handled more than 1 billion chat prompts.} These prompts share the same system prompts 
 that defines the functionality of Bing Chat. OpenAI introduces GPTs~\cite{gpts} to let users customize a ChatGPT for a specific purpose whose prompt template is the same across users. The commonality in prompts is crucial as it delineates the functionality and restrictions of LLM-based applications. The prompt structure in \autoref{fig:bing_prompt}\cite{bingprompt} includes a role definition, several examples to enhance the precision of LLM's behaviors and user query details. 
\chen{While the user input is dynamic, the task role is always fixed, and the few-shot examples could be quasi-static in that the same type of tasks use the same examples}. \chen{This is why} more than $94\%$ of prefix tokens could be repetitively used across LLM requests for various users (\autoref{tab:num_call}). Such commonality also exists in multi-agent applications. For example, MetaGPT~\cite{metagpt} and AutoGen~\cite{wu2023autogen} recurrently incorporate conversation history into the prompt over several rounds of LLM requests, leading to $72\%$ and $99\%$ redundancy respectively. These redundant sections excessively utilize GPU memory bandwidth and are computed for multiple times. 
\zh{Earlier results have proposed optimizations in LLM engines to avoid redundant GPU memory of shared prompt~\cite{vllm}. However, it is hard for public LLM services to} 
\zh{swiftly detect and co-locate the prompt-sharing requests, which be dynamically generated, from tons of diverse requests from diverse applications.} Without knowledge about the prompt structure, extensive token-by-token matching for every LLM request is expensive \zh{at the cluster level. Hence, if the cluster scheduler of public LLM service cannot dispatch prompt-sharing requests to the same engine, the engine-level redundancy avoidance optimizations would be hard to take effect}.

\begin{figure}[t]
    \centering
    \includegraphics[width=0.8\linewidth]{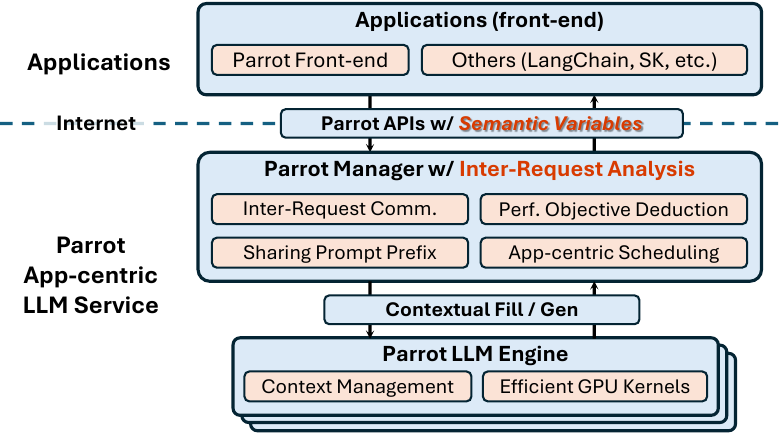}
    \caption{\sysname{} system overview.\label{fig:systemarch}}\vspace{-2mm}
\end{figure}

\section{\sysname{} Design}

\autoref{fig:systemarch} depicts the overview of \sysname{}'s design. \zh{\sysname{} provides a natural way of programming LLM applications with \sv{} annotations (\S\ref{sec:sv}), which is compatible of existing LLM orchestration frameworks, e.g., LangChain~\cite{langchain}. Centering on this abstraction, \sysname{} Manager is designed to schedule LLM requests at a cluster-level, by deriving the application-level knowledge (\S\ref{sec:primitives}) and optimizing end-to-end performance of application (\S\ref{sec:apps}).} \cf{ The manager will schedule the LLM requests to LLM \texttt{Engine}, which is formed by a GPU server (or a group of servers) in the cluster that can serve LLM requests independently.}  


\subsection{\sv{} \label{sec:sv}}

\zh{\sysname{} treats an LLM request as a semantic function\footnote{The term \emph{semantic function} is borrowed from Semantic Kernel~\cite{sk}.} implemented using natural language and executed by LLMs. 
\zh{A \sv{} is defined as a input or output variable of a semantic function, which is \cf{referred} as a placeholder in the prompt.} \autoref{fig:sv_code_test} shows a simplified example of  multi-agent application like MetaGPT~\cite{metagpt}. It contains two \texttt{SemanticFunction}s, one for the software engineer to write code and one for the QA engineer to write test code. It has three \sv{}s: \texttt{task}, \texttt{code}, and \texttt{test}, for task description, the code to be developed by the software engineer, and the test code to be developed by the QA engineer, respectively. Although existing LLM orchestration frameworks  (e.g., LangChain~\cite{langchain}) also allow placeholders in a prompt, however, the placeholders are rendered with real data before the submission, hence public LLM services cannot detect such a structure. Instead, \sysname{} relies on \sv{}s to preserve the prompt structure for further inter-request analysis in public LLM services side.} 



\begin{figure}[t]
    \centering
    \begin{minted}[
    frame=none,
    obeytabs=true,
    framesep=0mm,
    baselinestretch=0.9,
    fontsize=\footnotesize,
    xleftmargin=0em,
    breaklines,
    texcomments,
    escapeinside=||
    ]{python}
import Parrot as P
from Parrot.PerformanceCriteria import LATENCY

@P.SemanticFunction
def WritePythonCode(task: P.SemanticVariable):
""" You are an expert software engineer. 
    Write python code of {{input:task}}.
    Code: {{output:code}}
"""

@P.SemanticFunction
def WriteTestCode(
    task: P.SemanticVariable, 
    code: P.SemanticVariable):
""" You are an experienced QA engineer. 
    You write test code for {{input:task}}.
    Code: {{input:code}}.
    Your test code: {{output:test}}
"""

def WriteSnakeGame():
  task = P.SemanticVariable("a snake game")
  code = WritePythonCode(task)
  test = WriteTestCode(task, code)
  return code.get(perf=LATENCY), test.get(perf=LATENCY)
    \end{minted}
    \vspace{-0.3cm}
    \caption{Example: a multi-agent application in \sysname{}.\vspace{-0.3cm}}
    \label{fig:sv_code_test}
\end{figure}
\zh{In addition to the semantic functions, LLM application developers can further define orchestration functions that connect multiple semantic functions (e.g., \texttt{WriteSnakeGame} in \autoref{fig:sv_code_test}). The \sv{}s connecting multiple semantic functions form the data pipeline of multiple LLM requests in the public LLM service. }
A simple data flow analysis of the semantic functions can be done to reveals the connections of multiple LLM requests. 
E.g., in \autoref{fig:sv_code_test}, the \texttt{code} variable connects the two LLM requests originating from \texttt{WritePythonCode} and \texttt{WriteTestCode}, showing their sequential dependency. \cf{Different from traditional completion API, \sysname{} splits a completion request to \texttt{submit} operation and \texttt{get} operation (\S\ref{sec:impl}). \zh{A function calling of \texttt{SemanticFunction} will trigger the \texttt{submit} API to submit a LLM request with its prompt and input \sv{}s. The execution of a \texttt{SemanticFunction} is asynchronous thus it returns the futures of the output \sv{}s.} Through the \texttt{get} API, applications can fetch the value of an output \sv{} from the public LLM service in an on-demand manner. This asynchronous design allows \sysname{}-powered LLM service to receive all LLM requests not blocked by native functions and analyze their relationships just-in-time.}



\zh{The \texttt{get} operation supports annotation of performance criteria, showing the end-to-end performance requirement of an application, which can be end-to-end latency or throughput (extensible to more criteria like per-token latency when streaming, and time-to-first-token). For example, the final outputs, \texttt{code} and \texttt{test} in \autoref{fig:sv_code_test}, are fetched using \texttt{get} with an objective of end-to-end latency.} 
\cf{Criteria of middle variables will be automatically deduced and propagated from final outputs (\S\ref{sec:perfdeduct}). After propagation, each variable is attached to a criterion, which finally works by serving as a hint to Parrot's scheduler (\S\ref{sec:scheduling}).}  



\subsection{Primitives of Inter-Request Analysis\label{sec:primitives}}

\begin{figure}[t]
    \centering
    \includegraphics[width=1.0\linewidth]{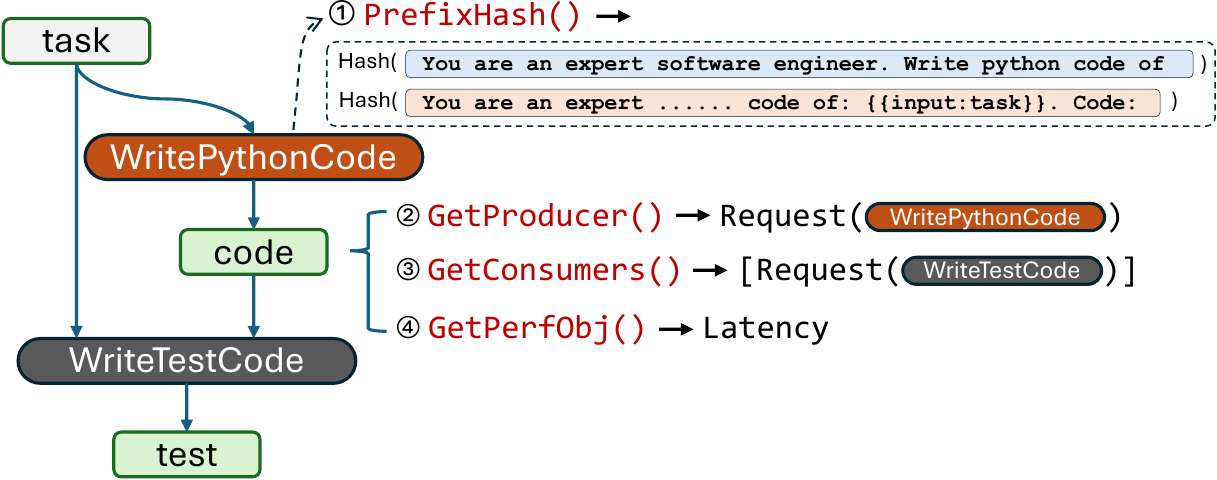}
    \caption{\zh{Primitives (selected) for Inter-Request Analysis.\vspace{-0.3cm}}}
    \label{fig:code_primitvies}
\end{figure}

\cf{In general, \sysname{} perform inter-request analysis mainly by two types of application-level information deduced from \sv{}: DAG of requests and prompt structure. \autoref{fig:code_primitvies} illustrates the DAG workflow of the example shown in \autoref{fig:sv_code_test} and the primitives used for inter-request analysis and optimizations.}

\zh{\paragraph{DAG-based analysis.} \cf{As requests, or \texttt{SemanticFunction}s, are submitted beforehand}
, \sysname{} can receive them all at once and analyze their correlations just-in-time on the service side. \cf{Parrot maintains a DAG-like data structure in each user's registered session. Each node is either a request or a \sv{} that connects different requests. When a request comes, \sysname{} inserts it to DAG by linking edges with \sv{}s it refers through placeholders in the prompts.}
\sysname{} can perform conventional dataflow analysis~\cite{nielson2015principles,tensorflow} using the primitives to get the producer and consumers of \sv{}s (i.e., \texttt{GetProducer} and \texttt{GetConsumers}) to recover dependency of LLM requests. Using the request DAG and the annotated performance criteria (via \texttt{GetPerfObj}) of final output \sv{}s, \sysname{} can deduct the request-level scheduling preference by analyzing the DAG and the performance objective of final outputs (\S\ref{sec:perfdeduct}).}

\zh{\paragraph{Prompt structure-based analysis.} Based on the prompt structure declared by \sv{}s, \sysname{} supports extracting the hash values of an LLM request at multiple positions split by \sv{}s (i.e., \texttt{PrefixHash}). For example, the prompt of \texttt{WritePythonCode} has two potential sharing prefix: the text before \texttt{\{\{input:task\}\}} and the text before \texttt{\{\{output:code\}\}}, thus there will be two prefix hash values generated. The prefix hashes of LLM requests will be used by swift detection of commonality across multiple requests, supporting both static and dynamically generated contents, as well as within the same type of  application or even across applications (\S\ref{sec:prefix}).}

\section{Optimizations with \sv{}} 
\label{sec:apps}

\subsection{Serving Dependent Requests}
\label{sec:irc}

\zh{To avoid the unnecessary client-side execution, it requires the dependency of requests at the application level, which is lost in today's public LLM services.} 
\cf{With the DAG and primitives illustrated in \S\ref{sec:primitives}, \sysname{} serves dependent requests efficiently through a graph-based executor. The executor polls constantly and sends it to corresponding engine once ready (i.e. producer requests are all finished), which allows instant execution and maximizes batching opportunities. For consecutive execution of dependent requests, materialized value is transmitted through a message queue allocated for corresponding \sv{}, avoiding unnecessary chatty communication between clients and LLM services.}

The value of a \sv{} in a request may require transformation before being exchanged, 
e.g., the value of a \sv{} is extracted from the JSON-formatted output of an LLM request, which is then fed into consecutive LLM requests. Similar to existing message queue systems that support message transformation (e.g., Kafka~\cite{kafka}), \sysname{} also supports string transformation to manipulate \sv{}s during value exchanging among LLM requests. \sysname{} supports most output parsing methods of LangChain~\cite{langchain}, which covers most use cases of LLM applications. 


\subsection{Performance Objective Deduction\label{sec:perfdeduct}}
\zh{To optimize the end-to-end performance of applications, we need to know the application-level performance criteria. To help deriving the request-level scheduling preference from the end-to-end application's performance requirement, we need to understand the workflow of the LLM application, which is the DAG of LLM requests derived by \sysname{}'s primitives. }

\zh{When an application annotates a \sv{} to prefer higher throughput, all requests generating this \sv{} (both directly or indirectly) will be marked as throughput-preferred when scheduling. This scheduling preference is usually beneficial for offline data processing, such as bulk document analysis.} 

\begin{figure}[t]
    \centering
    \includegraphics[width=0.9\linewidth]{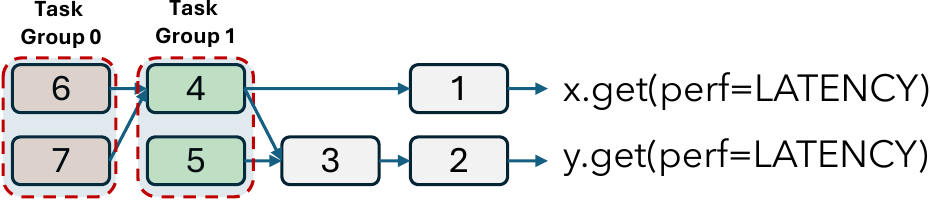}\vspace{-2mm}
    \caption{Performance deduction for an LLM-based application generating two latency-sensitive \sv{}.}\vspace{-2mm}
    \label{fig:deduce}
\end{figure}

Handling latency-sensitive \zh{applications} is more intricate. As demonstrated in \autoref{fig:mr_timeline}, achieving low end-to-end latency may sometimes require prioritizing throughput at the Mapping stage. 
The latency of individual requests can sacrificed so as to reduce the completion time of the entire DAG of requests. \sysname{} analyzes LLM requests in reverse topological order, beginning with those linked to latency-critical \sv{}, as depicted in \autoref{fig:deduce}. With the extracted DAG, LLM requests that directly result in latency-critical \sv{}s are labeled as latency-sensitive (Request 1 and 2), as are their immediate predecessors (Request 3). Parallel LLM requests at the same stage are grouped into a \emph{task group} (Task Groups 0 and 1). The scheduler should minimize the latency of the entire task group, \zh{often leading to a higher batch capacity for higher throughput of token generation}. 

\subsection{Sharing Prompt Prefix\label{sec:prefix}}

When an LLM request is scheduled to an LLM engine, a context on the engine is created to store the state of the model execution for this request (mainly KV cache). \zh{Existing works have proposed to share the KV cache of common prefix of prompts in LLM engines to save the GPU memory. However, as we have explained in \S\ref{sec:motivation}, today's public LLM service face diverse applications and requests, which is hard to identify the commonality at the cluster level. Token-by-token comparison is impractical due to high time complexity, especially for very long context with massive requests.} \zh{In \sysname{}, by exposing \sv{}s to LLM service, we can understand the prompt structure to automatically detect the commonality more efficiently at the granularity of \sv{}s. } Using \sysname{}'s primitive of \texttt{PrefixHash}, \sysname{} only needs to check the hash value at positions after each \sv{} in a request's prompt. \sysname{} maintains a key-value store, where each entry maps a (hashed) prefix of tokens to a list of requests, thus the scheduler can quickly check the opportunity in an online manner, \zh{supporting both static and dynamically-generated prompt within one application or even across different applications. }

\zh{Furthermore, we propose better GPU kernel for the attention computation of the requests with a common prefix. We first leverage vLLM's paged memory management~\cite{vllm} to save the redundent GPU memory. But vLLM's kernel still suffers from redundant computation and memory loading of the shared tokens. Therefore, we design a new Attention decoding algorithm by combining FlashAttenation~\cite{flashattention} and PagedAttention~\cite{vllm} that treat the shared and non-shared token separately. This significantly accelerates the attention of shared contexts (implementation details in \S\ref{sec:impl}).}



\subsection{Application-Centric Scheduling\label{sec:scheduling}}

\zh{To fix the problem of existing public LLM service that blindly optimize diverse individual requests, \sysname{}'s scheduling policy leverages the application-level knowledge to optimize the end-to-end performance.}
Specifically, the primary goal of \sysname{}'s scheduler is to meet the varied performance goals of LLM applications while optimizing GPU cluster utilization. As explained in \S\ref{sec:motivation}, a conflict arises when combining throughput and latency oriented requests: large batch sizes increase throughput and GPU efficiency but degrade latency, 
and vice versa. Transformer-based LLM inference 
is largely 
memory-bound, with latency influenced by the count of concurrent tokens within the engine. To meet performance targets of LLM applications, particularly latency, an LLM engine must regulate the token count below a specified threshold, which is determined by the LLM request with the most strict latency constraint. Therefore, \sysname{}'s scheduling principles are twofold: (1) group LLM requests with similar performance requirements to circumvent the conflict, and (2) maximize opportunities for sharing across requests.


\begin{algorithm}[t]
\caption{\sysname{}'s Request Scheduling.\label{alg:schedule}}

\KwData{Q: the request queue}
Q.sort() \Comment*[l]{Topological order} \label{line:sort}
\For{$r\in Q$}{
    SharedReqsInQueue, CtxInEngine = FindSharedPrefix($r$)\; \label{line:share1}
    \uIf{r.TaskGroup $\ne \varnothing$}{ \label{line:taskg1}
        $r^*$ = FindEngine(r.TaskGroup)\; \label{line:taskg2}
    }\uElseIf{SharedReqsInQueue $\ne \varnothing$}{ \label{line:share2}
        $r^*$ = FindEngine(SharedReqsInQueue)\;
    }\uElseIf{CtxInEngine $\ne \varnothing$}{
        $r^*$ = FindEngine(r, filter=CtxInEngine)\;\label{line:share3}
    }
    \If{$r^*$ = $\varnothing$}{ \label{line:other1}
        $r^*$ = FindEngine($r$)\;\label{line:other2}
    }
    Q.remove($r^*$)\; \label{line:remove}
}
\end{algorithm}

\autoref{alg:schedule} outlines the scheduling process of \sysname{}. With the extracted DAG, the system arranges the LLM requests according to their topological order (\autoref{line:sort}).  
\sysname{} tends to schedule requests belonging to the same application together to avoid the slowing down of interleaved scheduling (\S\ref{sec:eval_summary}). For requests identified as part of a task group through \sysname{}'s performance objective deduction, the scheduler attempts to allocate the entire task group together (\autoref{line:taskg1}-\autoref{line:taskg2}). Additionally, if \sysname{} detects other queued requests or running contexts with a common prefix, it tries to assign them to the same LLM engine (\autoref{line:share1}, \autoref{line:share2}-\autoref{line:share3}), to utilize \sysname{}'s context fork to reduce the redundant computation and GPU memory transactions. For an LLM request without the above opportunity, \sysname{} schedules the request independently (\autoref{line:other1}-\autoref{line:other2}). Due to limited space, we omit the details of how \sysname{} chooses LLM engines (i.e., \texttt{FindEngine}). Briefly, \sysname{} finds the engine that satisfies the scheduling preference of a request while minimizing the negative impacts.  For instance, if a latency-sensitive request is scheduled to an LLM engine that can run up to 64,000 tokens of throughput-driven requests, its capacity will be significantly reduced to 2,000 to satisfy its strict latency requirement. But, if it is scheduled to an engine that has already been running a latency-sensitive request, the capacity reduction is negligible.

\section{Discussion\label{sec:discussion}}

\paragraph{Dynamic Applications and Function Calling.} \zh{Currently, \sysname{} only supports cloud-side orchestration of LLM requests without involving dynamic control flow and native functions (e.g., Python Code). They still require client-side execution. We intentionally disable the offloading of these functions to public LLM services to minimize the security risks of malicious injection. For private LLM services whose LLM applications are trusted or there is a trusted zone to execute these functions, \sysname{}'s APIs can be easily extended with conditional connections and native code submission. Moreover, these extensions further enable new optimizations, e.g., we can speculatively pre-launch high-probability branches in dynamic applications based on past profiles. This also proves the potential of \sysname{}’s design when facing new types of applications. We leave these extensions as future works.}



\paragraph{Other Applications of Inter-Request Analysis.} The inter-request analysis in \sysname{} enables a new optimization space not limited to the ones we introduced in \S\ref{sec:apps}. 
A large-scale service has more scheduling features to consider, including handling outliers~\cite{outlier}, job failures~\cite{zaharia2008improving}, delay scheduling~\cite{delayscheduling}, fairness~\cite{altruistic,hived}, starvation~\cite{tiresias}, or supporting heterogeneous clusters~\cite{gavel,sia}, which have been widely studied in other systems. 
\sysname{} provides a new view from the perspective of LLM-based applications: we need to understand the interconnection and commonality of LLM requests to optimize applications' end-to-end performance. These features can be revisited in the LLM service system by considering the new characteristics of LLM applications. In this paper, we focus on \sysname{}'s mechanisms and a few use cases, leaving other optimizations as promising future works.

\paragraph{Parrot with LLM Orchestration Frameworks.} There have been several frameworks for developers to build LLM-based applications, e.g., LangChain~\cite{langchain}, SemanticKernel~\cite{sk}, and PromptFlow~\cite{promptflow}. The key function of these frameworks is to ``glue'' different LLM calls to accomplish a complex task (aka. LLM orchestration). \sysname{} can be integrated with these frameworks by extending their calling of LLM service APIs with \sv{}s. Most of these frameworks have already used a template-based approach in which developers can design a template with placeholders, and render the placeholders at runtime. These placeholders naturally have the same concept as \sysname{}'s \sv{}. However, because these frameworks will render the template prompt before the submission, LLM services lose the information on the prompt structure. 
To make these frameworks compatible with \sysname{}, both the template itself and the variables to render the template (using \sv{} in \sysname{}) need to be wrapped as a \texttt{SemanticFunction} so the necessary information is exposed to \sysname{}'s LLM service.

 \section{Implementation\label{sec:impl}}
\sysname{} is an end-to-end LLM service for LLM applications, implemented on Python with about 14,000 lines of code. Its front-end provides the abstraction of \sv{}, and \texttt{SemanticFunction}, which is transformed into \sysname{}'s APIs (implemented with FastAPI~\cite{fastapi}) to be submitted as LLM requests. A centralized \sysname{} manager handles the management of LLM requests, including \sv{}s, communication, and scheduling. We also build an LLM engine based on efficient kernels from vLLM~\cite{vllm}, xFormers~\cite{xFormers2022}, and ourselves. The engine supports advanced features for LLM serving, including paged memory management~\cite{vllm} and continues batching~\cite{orca}. 
\sysname{}'s front-end and manager are implemented in 1,600 and 3,200 lines of Python, respectively. \sysname{}'s LLM engine is implemented in 5,400 lines of Python and 1,600 lines of CUDA. We have implemented OPT~\cite{opt} and LLaMA~\cite{llama} with  PyTorch~\cite{pytorch} and Transformers~\cite{transformers}. 

\paragraph{APIs.} \cf{Applications programmed by \texttt{SemanticFunction}s or other frontends are finally lowered to requests to universal APIs through different adapters. \sysname{} provides OpenAI-like APIs with the extension of \sv{}s. The request body of two operations mentioned in \S\ref{sec:sv} is shown as follows:}

\begin{minted}[
    frame=none,
    obeytabs=true,
    framesep=0mm,
    baselinestretch=0.9,
   % fontsize=\footnotesize,
    fontsize=\small,
    xleftmargin=0em,
    escapeinside=||,
    breaklines,
    texcomments,
    ]{python}
(submit) {"prompt": str, "placeholders": [{"name": str, "in_out": bool, "semantic_var_id": str, "transforms": str}, ...], "session_id": str}

(get) {"semantic_var_id": str, "criteria": str, "session_id": str}
    \end{minted} 

 

In addition to the static string prompt, \sysname{} preserves the input and output placeholders. A placeholder is associated with a semantic variable either for rendering the input or parsing the output. As introduced in \S\ref{sec:irc}. \sysname{} supports transformations before the input or after the output. \sysname{} also supports other APIs for setting and fetching the value of \sv{}s. The error message will be returned when fetching an \sv{}, whose intermediate steps fail (including engine, communication, and string transformation).

\paragraph{Kernel Optimization.} vLLM's GPU kernel, while capable of reusing results cached in GPU memory for shared prefix tokens in a prompt, sometimes excessively 
reloads these tokens from global to shared memory, impeding attention score computations. Using OpenAI Triton~\cite{triton} and CUDA, we have developed a novel GPU kernel, integrating concepts from PagedAttention~\cite{vllm} and FlashAttention~\cite{flashattention,dao2023flashattention}, to accelerate attention decoding computation involving shared prefixes. This kernel retains PagedAttention's approach of storing the key-value (KV) cache in disparate memory segments and utilizes a page table per request to monitor block status and placement. 
Furthermore, employing FlashAttention principles, the kernel maximizes data reuse within shared memory. Unlike reloading tiles repeatedly in the PagedAttention's implementation, it loads KV cache tiles for the shared prefix to shared memory only once, diminishing memory transactions between the L2 Cache and Shared Memory. 
The kernel initially calculates interim attention metrics (including attention scores, \texttt{qk\_max}, \texttt{exp\_sum}) for the shared prefix using the loaded tiles and records these back to HBM. Subsequently, it processes the new tokens' partial attention beyond the prefix, amalgamating this with the prefix's interim results to derive the ultimate attention output. 

\paragraph{Universal Engine Abstraction.} \cf{\sysname{}'s cluster manager controls multiple engines running various models, tokenizers, KV cache layouts, etc. To enable \sysname{}'s optimizations, LLM engines need to support (1) stateful generation (e.g., guidance~\cite{guidance}) and (2) sharing KV cache states across different requests.  
Hence we propose a universal abstraction to describe the minimal capability required to LLM engines to be integrated into \sysname{}.} 
\begin{minted}[
    frame=none,
    obeytabs=true,
    framesep=0mm,
    baselinestretch=0.9,
   % fontsize=\footnotesize,
    fontsize=\small,
    xleftmargin=0em,
    escapeinside=||,
    breaklines,
    texcomments,
    ]{python}
def Fill(token_ids: List[int], context_id: int, parent_context_id: int)
def Generate(sampling_configs: Dict, context_id: int, parent_context_id: int)
def FreeContext(context_id: int)
    \end{minted} 
\cf{These three methods not only cover the basic completion functionality of LLM inference engine, but also provide a flexible context management interface. The \texttt{Fill} method processes the initial prompt tokens, calculates and fills the KV cache into corresponding context. The \texttt{Generate} method produces tokens via generative decoding that produces one token per iteration until it reaches the length limit, user-defined termination character or EOS (end-of-sequence) token, under certain sampling configurations (e.g. temperature). \texttt{Fill}s and \texttt{Generate}s are scheduled and batched by engine's scheduler per iteration using continuous batching~\cite{orca}. Creating and forking contexts can also be realized with these two methods by setting \texttt{context\_id} and \texttt{parent\_context\_id}, respectively. The \texttt{FreeContext} method explicitly frees a context (i.e. free its KV cache in GPU memory). Separating \texttt{Fill} and \texttt{Generate} not only fits \sv{} naturally: constant text and input values are processed by \texttt{Fill}; the output values are generated by \texttt{Generate}, but also breaks the request-level dependency into a finer granularity, enabling more parallel execution opportunities\zh{~\cite{zhong2024distserve,patel2023splitwise,agrawal2024taming,holmes2024deepspeed}}.
}

\begin{table}[b]
\centering
\scriptsize{
\begin{tabular}{l|c|c|c|c}
\hline
Workload                                                     & \begin{tabular}[c]{@{}c@{}} Serving\\ Dependent\\ Requests.\end{tabular} & \begin{tabular}[c]{@{}c@{}}Perf. Obj.\\ Deduction\end{tabular} & \begin{tabular}[c]{@{}c@{}}Sharing \\ Prompt\end{tabular} & \begin{tabular}[c]{@{}c@{}}App-centric \\ Scheduling\end{tabular} \\ \hline
Data Analytics                                               & \checkmark                                         & \checkmark                                        &                                                         & \checkmark                                           \\ \hline
\begin{tabular}[c]{@{}l@{}}Serving Popular\\ LLM Applications\end{tabular} &                                                                 &                                                                & \checkmark                                 & \checkmark                                           \\ \hline
Multi-agent App.                                             & \checkmark                                         & \checkmark                                        & \checkmark                                 & \checkmark                                           \\ \hline
Mixed Workloads                                              & \checkmark                                         & \checkmark                                        &                                                         & \checkmark                                           \\ \hline
\end{tabular}}
\caption{The workloads and the optimizations taking effect.\label{tab:workload}}
\end{table}
\section{Evaluation}
\label{sec:eval}

\subsection{Experimental Setup}
\paragraph{Testbed.} We evaluate \sysname{} with two separate setups for single-GPU and multi-GPU experiments. The single-GPU evaluations use a server with a 24-core AMD-EPYC-7V13 CPUs equipped with one NVIDIA A100 (80GB) GPU. The multi-GPU evaluations use a server with 64-core EPYC AMD CPU and four NVIDIA A6000 (48GB) GPUs. Both servers run CUDA 12.1 and cuDNN 8.9.2.

\paragraph{Workloads.} Our evaluations are performed to run four representative LLM applications. Each LLM engine uses one GPU and runs a LLaMA 13B or LLaMA 7B model~\cite{llama} . For LLM-based data analytics on long documents, we use the Arxiv dataset~\cite{arxiv-march},  executing chain and map-reduce summarizations on an extensive collection of academic papers.  To investigate the sharing opportunities of LLM-based applications with many users, we run the prompts from Bing Copilot and GPTs~\cite{gpts} with synthesized user queries. For multi-agent applications, we build a multi-agent programming application using MetaGPT~\cite{metagpt}, which contains a system architect to design APIs, multiple programmers to write code for different files, reviewers to share review comments. The programmers will also revise the code based on comments. For chat service workloads, we derived scenarios from the ShareGPT dataset~\cite{sharegpt}, which mirrors real LLM chat conversations.  According to the distribution of our measurement, we introduced a random delay of $200\sim 300$ ms to LLM requests to emulate typical network overhead seen over the Internet. To create realistic workloads, we documented the LLM responses using GPT-4~\cite{gpt4}, ensuring the LLaMA models generated text of similar length for system performance analysis. \autoref{tab:workload} presents the workloads and their optimizations in \sysname{}.  

\begin{figure}[t]
	\centering
    \begin{subfigure}{0.49\linewidth}
        \centering
        \includegraphics[width=1\linewidth]{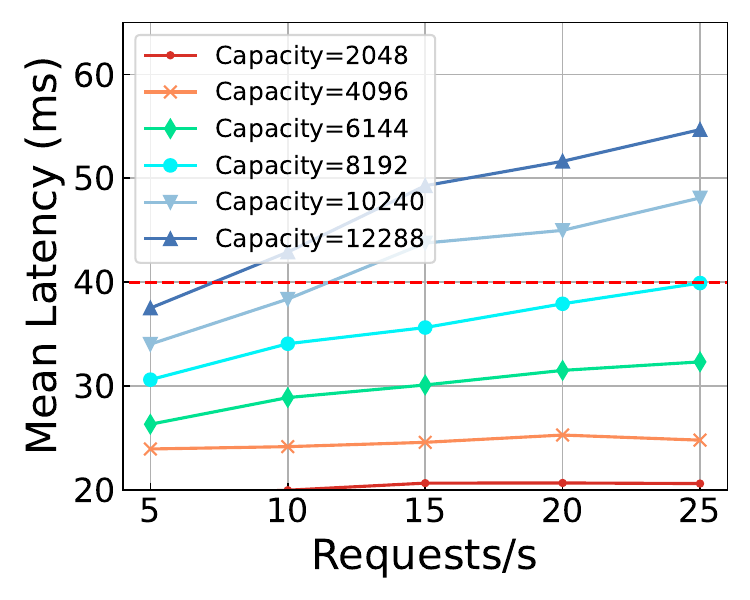}\\\vspace{-2mm}
        \caption{Mean Latency}
    \end{subfigure}~~
    \begin{subfigure}{0.49\linewidth}
        \centering
        \includegraphics[width=1\linewidth]{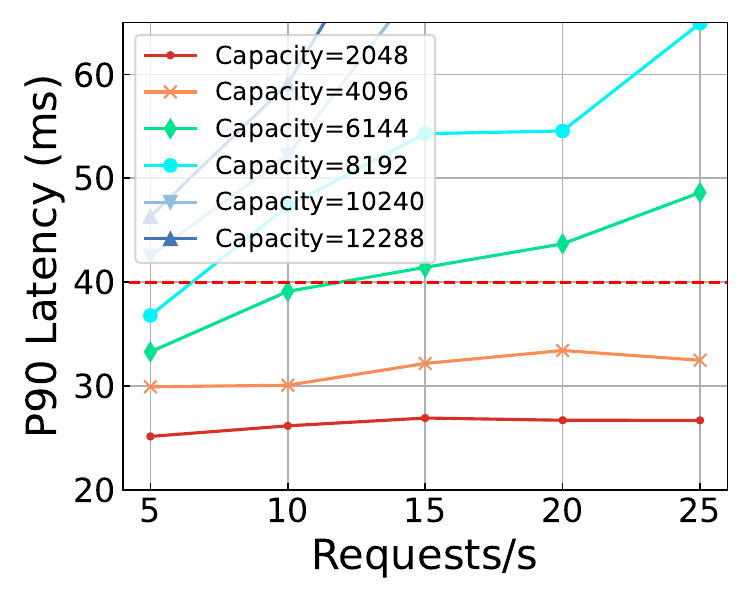}\\\vspace{-2mm}
        \caption{P90 Latency}
    \end{subfigure}
    \caption{Latency (per output token) of vLLM with varying token capacities and request rates. Requests are sampled from ShareGPT~\cite{sharegpt} and their arrival time follows Poisson distributions\label{fig:vio}.}\vspace{-2mm}
\end{figure}
\paragraph{Baseline.} We benchmark \sysname{} against sate-of-the-art solutions for building LLM applications and serving LLM requests. The majority of LLM applications used in our baseline comparisons are developed using LangChain~\cite{langchain}, which is the predominant framework for LLM application development. The LLM applications in baselines leverage OpenAI-style chat completion APIs as provided by FastChat~\cite{fastchat}. FastChat is a widely recognized open-source LLM serving system with over 30,000 stars on its repository. Incoming requests to FastChat are allocated to LLM engines that run either HuggingFace's Transformers library~\cite{transformers} or vLLM~\cite{vllm}, both of which incorporate cutting-edge enhancements for LLM execution, such as FlashAttention~\cite{flashattention}, PagedAttention~\cite{vllm}, and continuous batching techniques~\cite{orca}. The default scheduling strategy employed by FastChat assigns incoming requests to the LLM engine with the smallest current queue. Since existing LLM services typically expose their functionality through "chat" completion APIs, baseline assessments treat all requests as independent and assume a high sensitivity to latency. To manage token generation response times, each LLM engine is subject to a capacity threshold, which is the aggregate token count from all active requests on the engine. 

\zh{Since existing LLM token generation is usually bound by memory bandwidth, the per-token generation latency of an engine is mainly affected by the number of running tokens in a batch.} 
As depicted in \autoref{fig:vio}, our experiments indicate that 
\cf{the latency per output token, i.e. TPOT (Time-per-output-token)} for vLLM, with continuous batching enabled, experiences a notable uptick when the engine's workload \zh{using a batch capacity beyond 6144. In our evaluation, we use the setting that an LLM engine can keep its generation latency under 40 ms/s \cf{for latency-sensitive requests}, consistent with our experience of OpenAI's LLM services. } 
When all LLM engines hit their maximum capacity, any additional LLM requests are queued in a FIFO (First In, First Out) manner, awaiting the completion and release of resources by ongoing tasks. \zh{Serving longer context (e.g., 32k or even 1M tokens) within a satisfactory latency require either more GPUs using tensor-parallel~\cite{megatron} or sequence-parallel~\cite{colossalai} approaches, or approximate attention (e.g., StreamingLLM~\cite{xiao2023streamingllm}), which is beyond the scope of this paper.} 

\subsection{Data Analytics on Long Documents~\label{sec:eval_summary}}
Our experimental analysis within data analytics randomly picks ten long documents from the Arxiv-March dataset~\cite{arxiv-march}, using chain-summary and map-reduce summary. Each document has over 20,000 tokens. The results measures the mean end-to-end latency across all documents. 

\begin{figure}[t]
	\centering
    \begin{subfigure}{0.49\linewidth}
        \centering
        \includegraphics[width=1\linewidth]{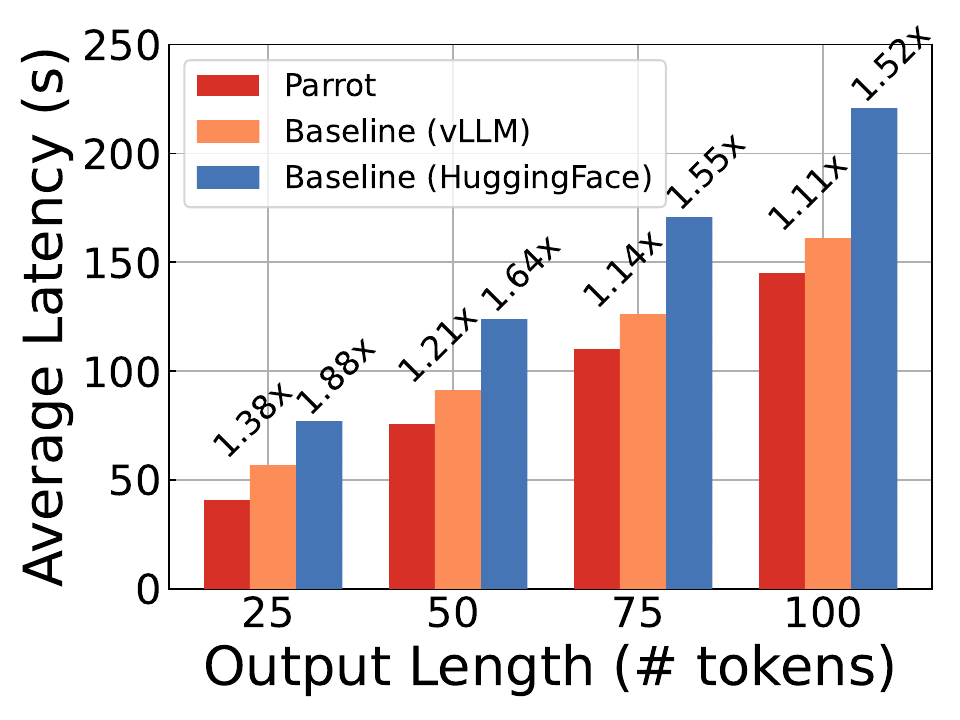}\\\vspace{-2mm}
        \caption{Output lengths\label{fig:expt_chain_single_olen}}
    \end{subfigure}~~
    \begin{subfigure}{0.49\linewidth}
        \centering
        \includegraphics[width=1\linewidth]{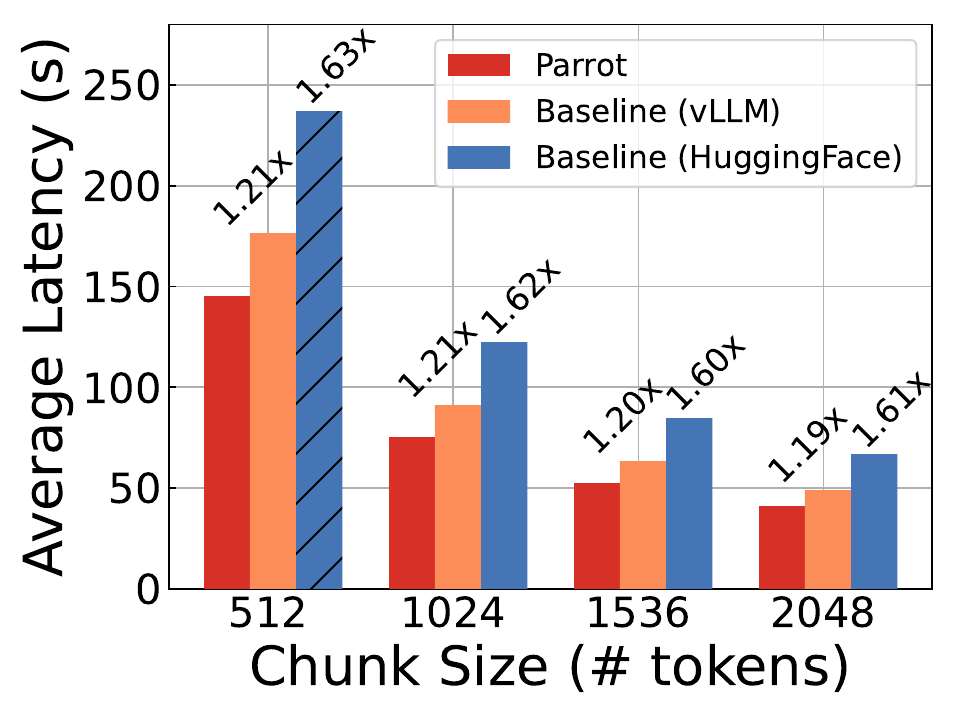}\\\vspace{-2mm}
        \caption{Chunk sizes\label{fig:expt_chain_single_csize}}
    \end{subfigure}
    \caption{Average E2E latency of chain summarization with varying output lengths and chunk sizes.\label{fig:expt_chain_single}}\vspace{-2mm}
\end{figure}

\paragraph{Chain-style Applications.} Our evaluation demonstrates how \sysname{} enhances chain summarization by mitigating the excessive communication overhead stemming from client interactions. \autoref{fig:expt_chain_single} presents the average end-to-end latency for summarizing a single document using one LLM engine (A100, LLaMA 13B) . We adjust the chunk size (the count of tokens per chunk) and the output length, with results shown in \autoref{fig:expt_chain_single_olen} and \autoref{fig:expt_chain_single_csize}, respectively. \sysname{} achieves a reduction in end-to-end latency by as much as $1.38\times$ and \cf{$1.88\times$} compared to the baselines employing vLLM and HuggingFace, respectively. The efficiency of \sysname{} primarily stems from the decreased network latency, which is a consequence of reduced client interaction. As the output length increases, the time spent on generation becomes more significant, leading to a diminishing advantage for \sysname{} over the baseline. By increasing the chunk size, we decrease the number of chunks, yet the extent of the speedup is contingent upon the network latency savings for each chunk. Given that token generation is substantially more time-consuming than prompt processing, we observe a consistent speedup with variable chunk sizes and a fixed output length ($1.2\times$ and $1.66\times$ relative to vLLM and HuggingFace, respectively). This indicates that \sysname{}'s optimization for dependent LLM requests is particularly beneficial for shorter outputs, which are prevalent in various LLM applications such as summarization, short answer generation, scoring, and choice provision. Due to HuggingFace's slower performance relative to vLLM, subsequent evaluations focus solely on the comparison between \sysname{} and vLLM.  

\begin{figure}[t]
	\centering
    \begin{subfigure}{0.49\linewidth}
        \centering
        \includegraphics[width=1\linewidth]{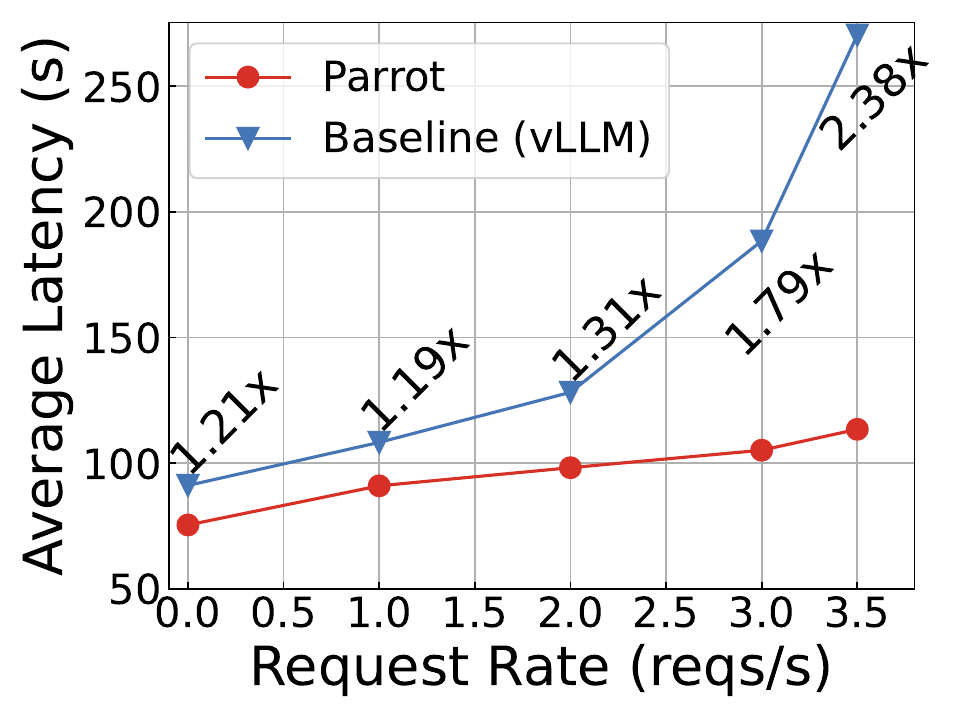}\\\vspace{-1mm}
        \caption{With background requests\label{fig:expt_chain_multi_bkg}}
    \end{subfigure}~~
    \begin{subfigure}{0.49\linewidth}
        \centering
        \includegraphics[width=1\linewidth]{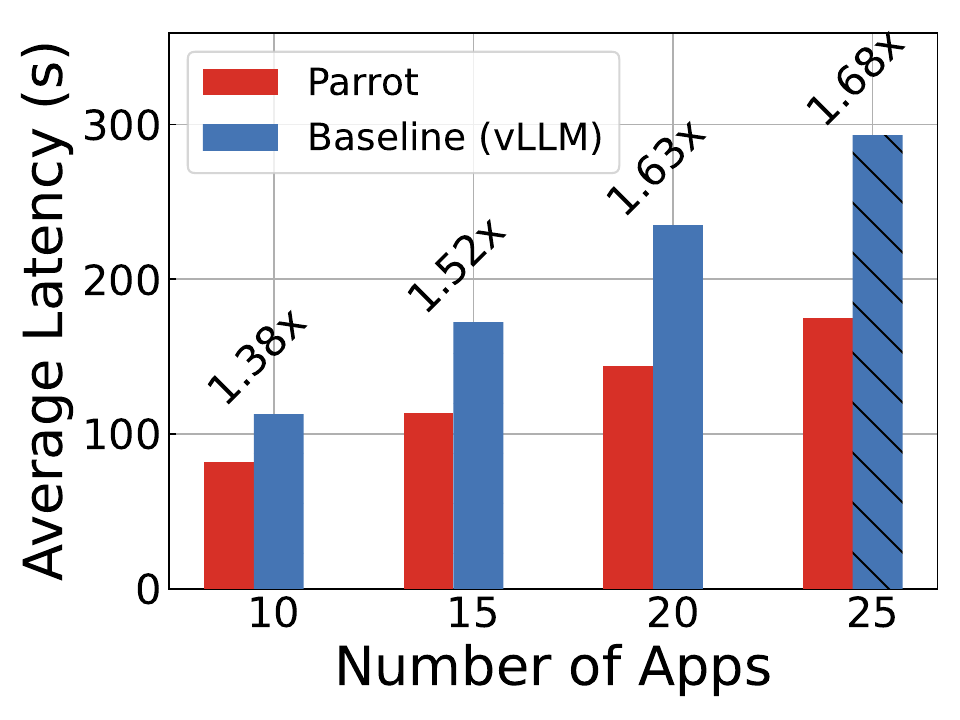}\\\vspace{-1mm}
        \caption{Multiple summary apps.\label{fig:expt_chain_multi_app}}
    \end{subfigure}
    \caption{Average \cf{E2E} latency of chain-summary with background requests or other chain-summary applications.\label{fig:expt_chain_bkg_multi}}
\end{figure}
\begin{figure}[t]
    \centering
    \includegraphics[width=1\linewidth]{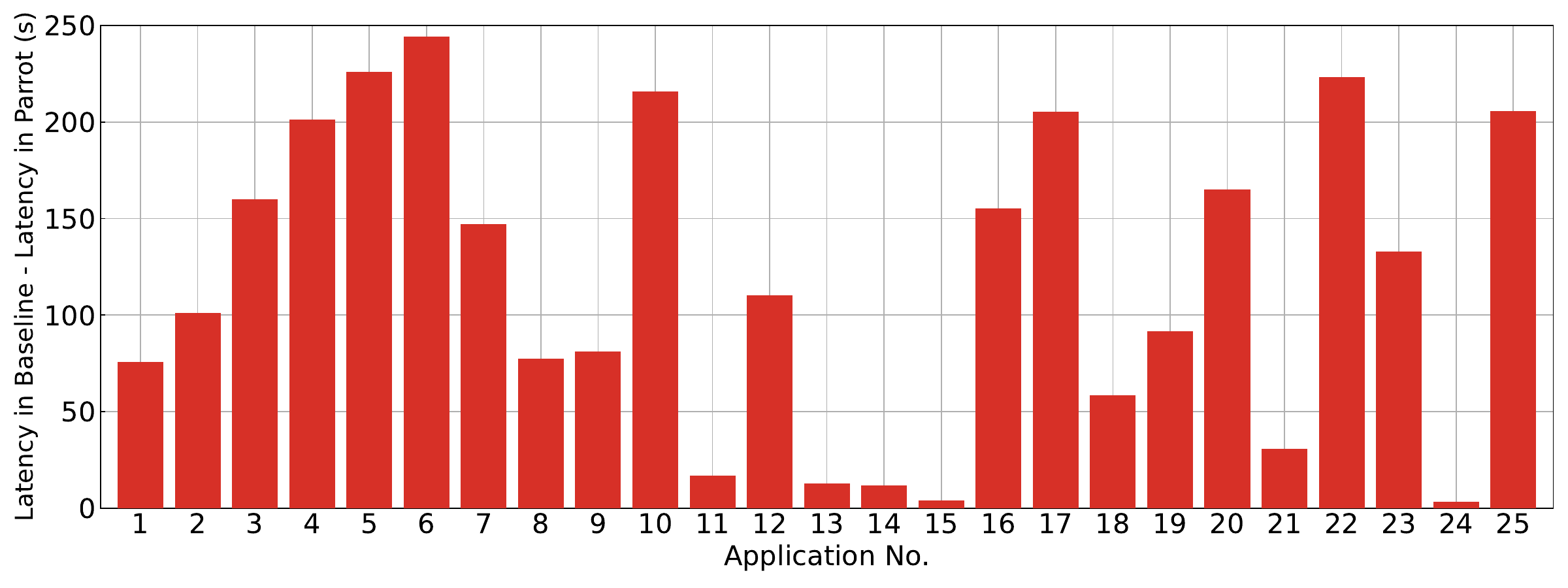}
    \caption{\cfnew{The difference in E2E latency of the 25 chain-summary application between Baseline and Parrot. All applications finish earlier in Parrot.} \label{fig:expt_chain_multi_jct_diff}}
\end{figure}

\autoref{fig:expt_chain_multi_bkg} extends the evaluation by introducing background LLM requests at varying rates to examine the capability of \sysname{} in mitigating additional queuing delays for dependent requests. \sysname{} slashes the end-to-end latency by a factor of $2.38\times$ in comparison to the baseline (vLLM). With \sysname{}, as soon as the summary for the first chunk is completed, the subsequent chunk is processed immediately by incorporating the summaries of previous chunks into the prompt, which aids in generating the summary for the next chunk. In contrast, the baseline treats all LLM requests individually. As a result, in addition to the network latency from client interactions, subsequent requests must re-enter the queue, leading to added queuing delays.   \autoref{fig:expt_chain_multi_app} further illustrates the end-to-end latency when multiple chain-summary applications are submitted concurrently, with each application tasked with generating a summary for a separate document. \sysname{} manages to reduce the average end-to-end latency for all applications by \cfnew{$1.68\times$} without slowing down any applications compared to the baseline \cfnew{according to \autoref{fig:expt_chain_multi_jct_diff}.} The baseline, by interleaving the execution of different applications, exacerbates the slowdown of the end-to-end latency for all applications. These experiments validate that recognizing the interconnections of LLM requests can significantly enhance end-to-end performance, as opposed to processing requests in isolation.  


\begin{figure}[t]
	\centering
    \begin{subfigure}{0.49\linewidth}
        \centering
        \includegraphics[width=1\linewidth]{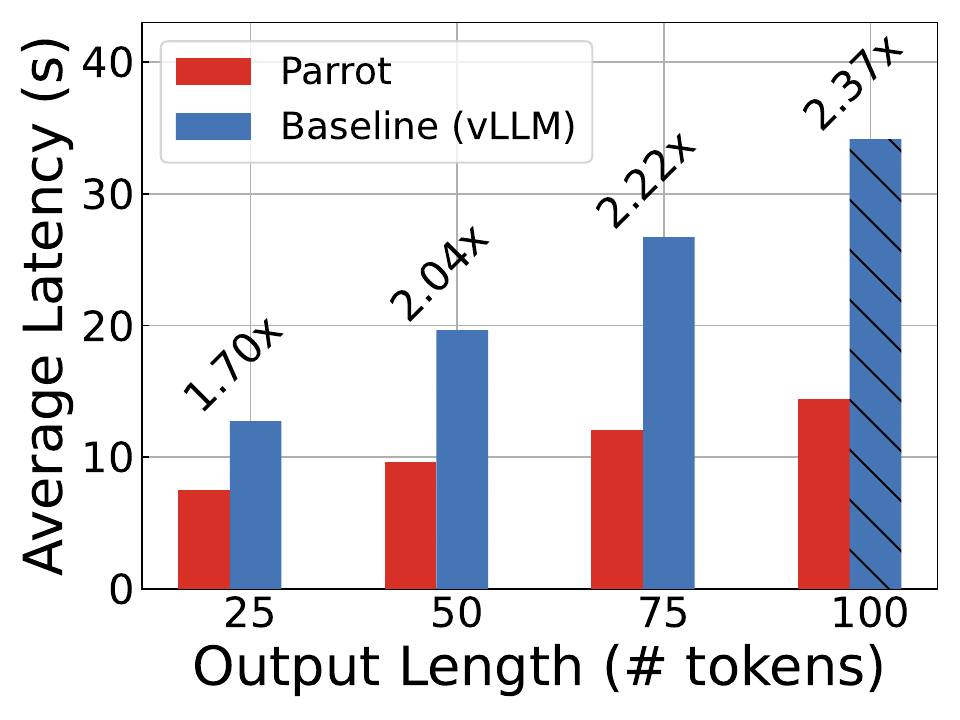}\\\vspace{-1mm}
        \caption{Output lengths\label{fig:expt_mr_olen}}
    \end{subfigure}~~
    \begin{subfigure}{0.49\linewidth}
        \centering
        \includegraphics[width=1\linewidth]{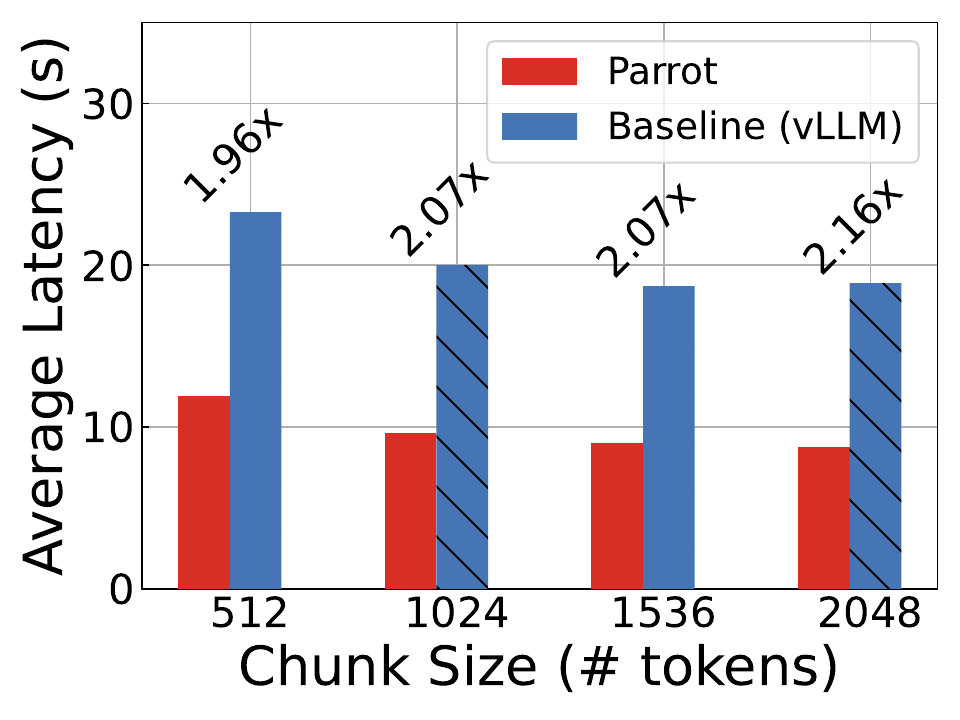}\\\vspace{-1mm}
        \caption{Chunk sizes\label{fig:expt_mr_csize}}
    \end{subfigure}
    \caption{Average \cf{E2E} latency of Map-Reduce document summary with varying output lengths and chunk sizes.\label{fig:expt_mr}}
\end{figure}

\paragraph{Map-Reduce Applications.} An alternative implementation of the document summarization application follows the map-reduce paradigm as depicted in \autoref{fig:example_map_reduce}. This approach consists of multiple parallel mapping LLM requests, where each request summarizes a distinct segment of the document, followed by a reducing LLM request that aggregates these individual summaries into a final summary. As shown in \autoref{fig:expt_mr}, \sysname{} realizes a $2.37\times$ acceleration over the baseline with one LLM engine (A100, LLaMA 13B). Since the mapping LLM requests are independent, they are dispatched concurrently by both \sysname{} and the baseline. The primary advantage of \sysname{} stems from its deduction of a performance objective that identifies the mapping tasks as a task group. By recognizing this relationship, \sysname{} is capable of optimizing the latency of the entire task group through larger batch sizes, which in turn enhances throughput. In contrast, the baseline processes each LLM request in isolation, operating under the presumption that they are all sensitive to latency. This constrains the baseline to utilize a limited token capacity (4096 tokens) on the LLM engine to achieve optimal latency for individual tasks, which is detrimental to the end-to-end performance of applications. It underscores the necessity for LLM services to distinguish LLM requests to optimize the end-to-end performance of varied LLM applications.  


\subsection{Serving Popular LLM Applications\label{sec:eval_many}}
\begin{figure}[t]
    \centering
    \includegraphics[width=0.9\linewidth]{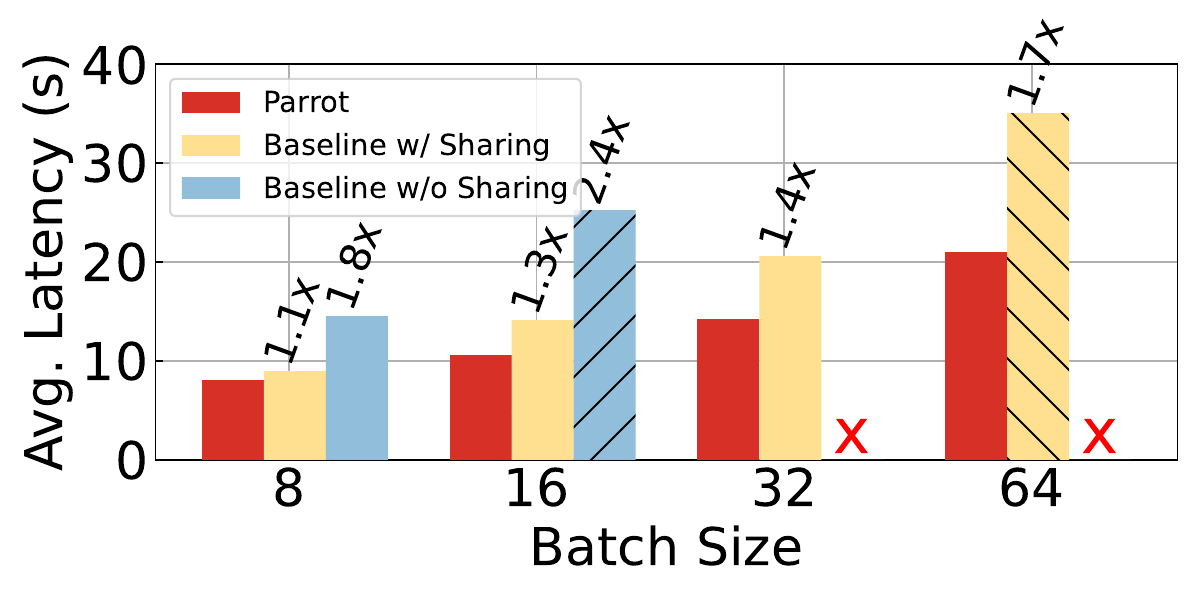}\vspace{-0.3cm}
    \caption{Latency of Bing Copilot with varying batch sizes.}\vspace{-0.3cm}
    \label{fig:expt_bing}
\end{figure}

\begin{figure}[t]
	\centering
    \begin{subfigure}{0.49\linewidth}
        \centering
        \includegraphics[width=1\linewidth]{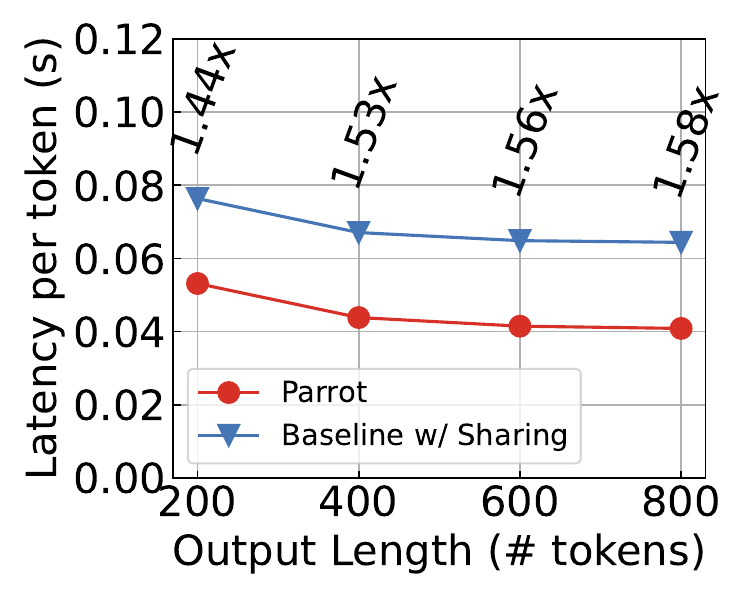}\\\vspace{-1mm}
        \caption{Batch Size = 32}
    \end{subfigure}~~
    \begin{subfigure}{0.49\linewidth}
        \centering
        \includegraphics[width=1\linewidth]{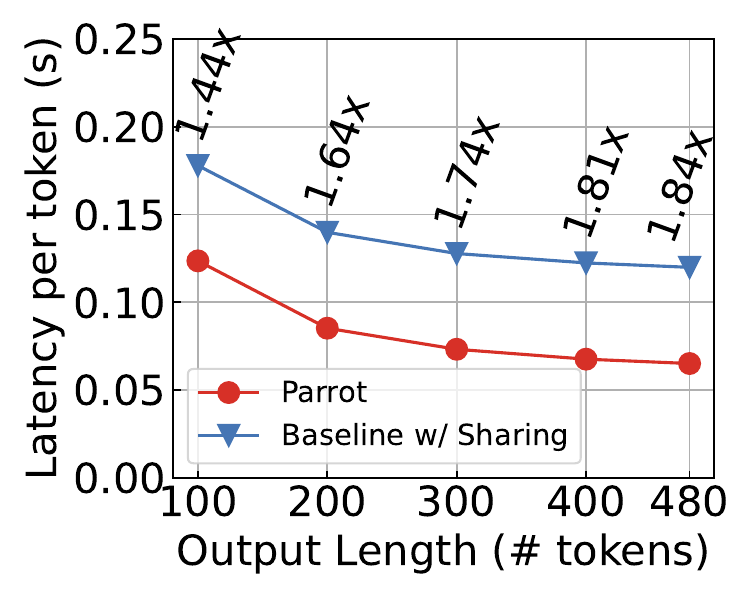}\\\vspace{-1mm}
        \caption{Batch Size = 64}
    \end{subfigure}
    \caption{Latency per \cf{output} token of Bing Copilot.\label{fig:expt_bing_bs}}\vspace{-0.3cm}
\end{figure}

Production applications need to face massive users. As explained in \autoref{fig:bing_prompt}, developers often need to use a very long system prompt to define the behavior of LLMs. Therefore, users of the same LLM application often use the shared prompt, which can benefit from \sysname{}'s context fork mechanism and \sysname{}'s scheduling policy that co-locates LLM requests sharing a long prompt prefix. Because we do not have access to the intermediate steps of Bing Copilot, we only evaluate the final request generating the response to users. We synthesized 64 requests from the length distribution we measured using Bing Copilot. The system prompt length is about 6000 tokens. The output lengths ranges from 180 to 800 tokens. \autoref{fig:expt_bing} shows the average request latency of Bing Copilot of \sysname{} and the baselines. Because the LLM service in the baseline system does not know the prompt structure, it is hard to infer the shared prompt from massive LLM requests. Compared to the baseline without sharing prompt, \sysname{} achieves $1.8\times\sim 2.4\times$ speedup for batch sizes of 8 and 16. Further increasing the batch size leads to out-of-memory due to the massive KV cache of shared system prompt. We also build an advanced baseline using vLLM's paged attention to support sharing the prompt with a static prefix. Both \sysname{} and vLLM use the paged memory management~\cite{vllm}, thus both systems can hold the same number of tokens in an LLM engine (A100, LLaMA 7B). \sysname{} further achieves $1.1\times\sim 1.7\times$ speedup over vLLM because of the better GPU kernel. Although vLLM can save extra memory usage of the shared prompt, its GPU kernel still has to reload the tokens repeatedly. Given that the token generation of LLMs is bound by memory bandwidth, such redundant memory loading slows down the end-to-end inference. By combining FlashAttention and PagedAttention, \sysname{} only needs to load the tokens of the shared prompt once, when computing the attention from the diverged tokens of different users. \sysname{}'s speedup of shared prompt mainly comes from the token generation, thus the longer output length leads to higher improvement. \autoref{fig:expt_bing_bs} shows \sysname{} achieves $1.58\times$ and $1.84\times$ speedup compared to vLLM using paged attention, showing $40~ms$ per-\cf{output-}token latency at a batch size of 32. 

\begin{figure}[t]
    \centering
    \includegraphics[width=0.9\linewidth]{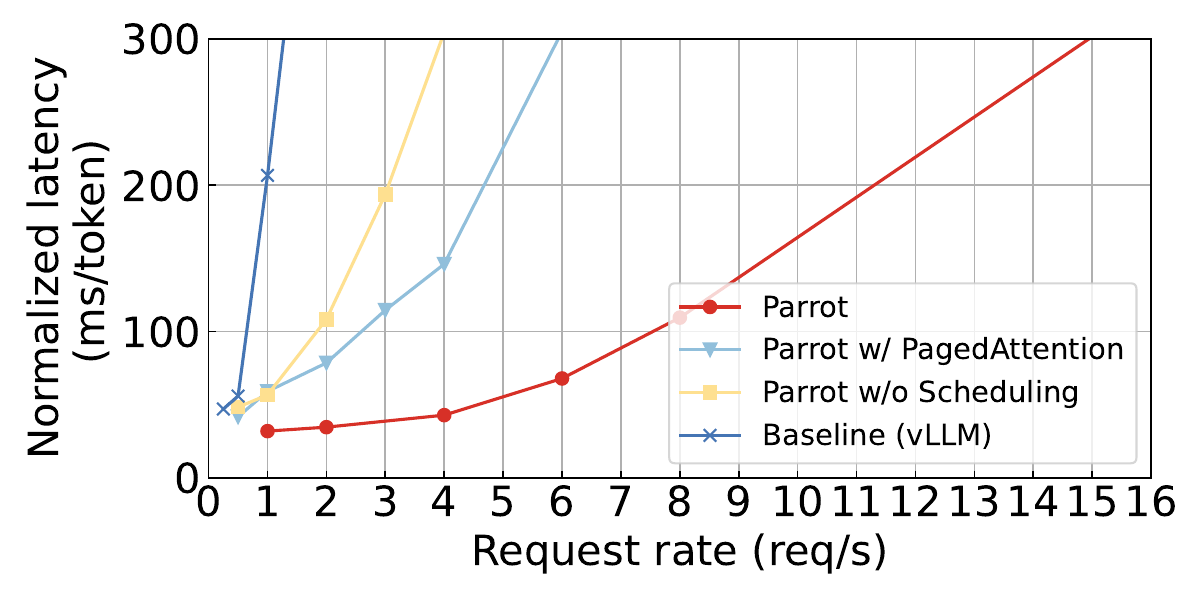}
    \caption{\cfnew{Serving multiple GPTs applications.} \label{fig:expt_gpts}}
\end{figure}

In \autoref{fig:expt_gpts}, we further evaluated the serving of multiple GPTs applications~\cite{gpts}, each of which has multiple users, in a multi-GPU cluster. Four A6000 (48GB) GPUs are deployed with four LLM engines (LLaMA 7B). We select four GPTs applications in four popular categories including productivity, programming, image generation, and data analysis. The LLM requests are randomly generated from the four categories with equal probability. LLM requests arrive at fixed rates following Poisson distribution. \sysname{} can sustain $12\times$ higher request rates compared to the baseline without sharing. Because the baseline's scheduling policy is not aware of the shared prompt within each LLM application, the requests are mixed in all LLM engines making it impossible to reuse the common prompt prefix. \sysname{}'s scheduling policy co-locates LLM requests of the same applications to maximize the sharing opportunity, achieving both lower inference latency and higher cluster throughput. \cfnew{After turning off such affinity scheduling policy, \sysname{} only exhibits $3\times$ higher request rates compared to the baseline, because the requests with shared prefix are often dispatched to different engines thus reduced the sharing opportunities.} Moreover, \sysname{}'s attention kernel helps \sysname{} to achieve $2.4\times$ higher rate compared to \sysname{} using vLLM's PagedAttention, by avoiding the redundant memory loading for attention of shared prompts.


\subsection{Multi-agent Applications}
\begin{figure}[t]
	\centering
    \begin{subfigure}{0.9\linewidth}
        \centering
        \includegraphics[width=1\linewidth]{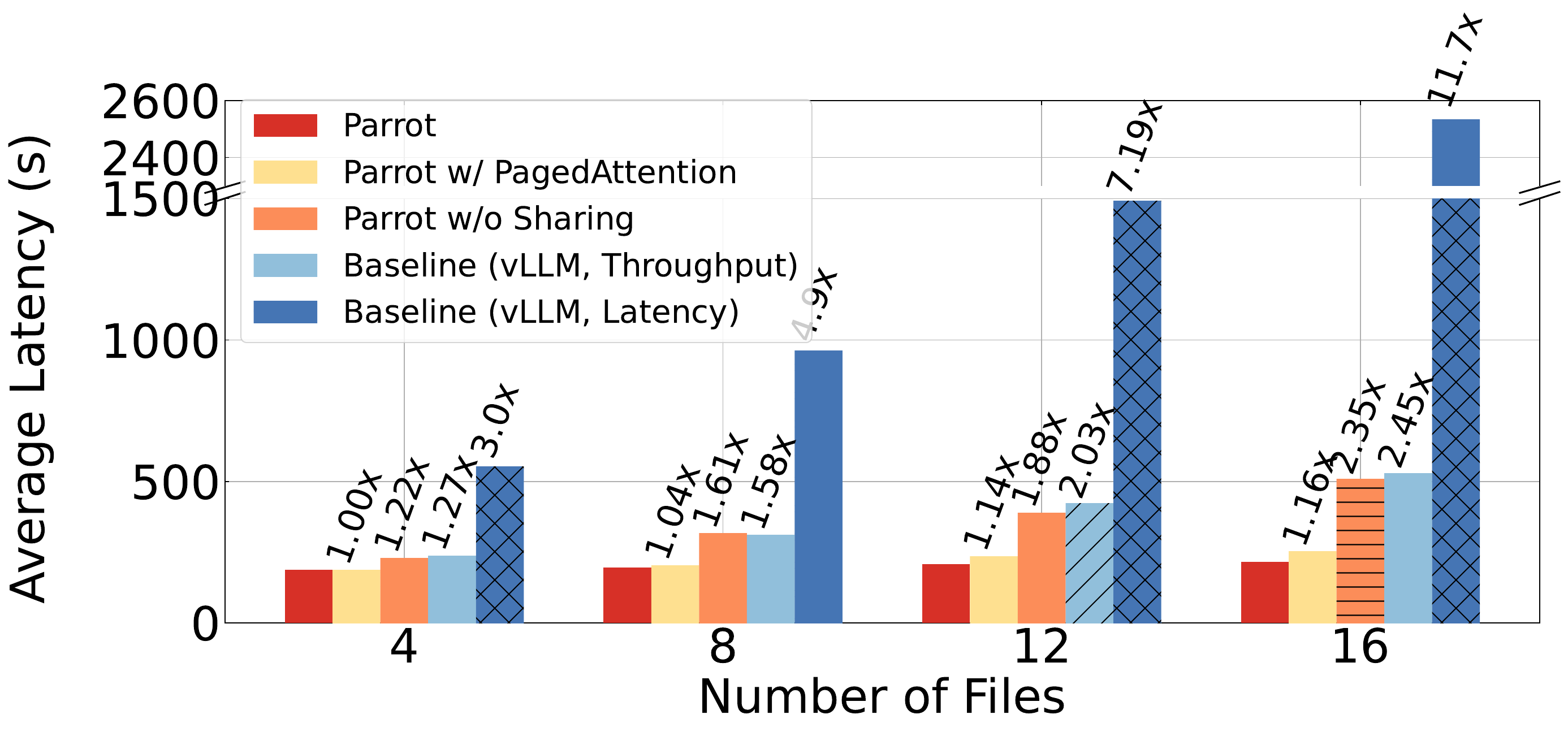}\\\vspace{-1mm}
        \caption{\cfnew{End-to-end Latency} \label{fig:expt_multi_latency}}
    \end{subfigure}\\
    \begin{subfigure}{0.87\linewidth}
        \centering
        \includegraphics[width=1\linewidth]{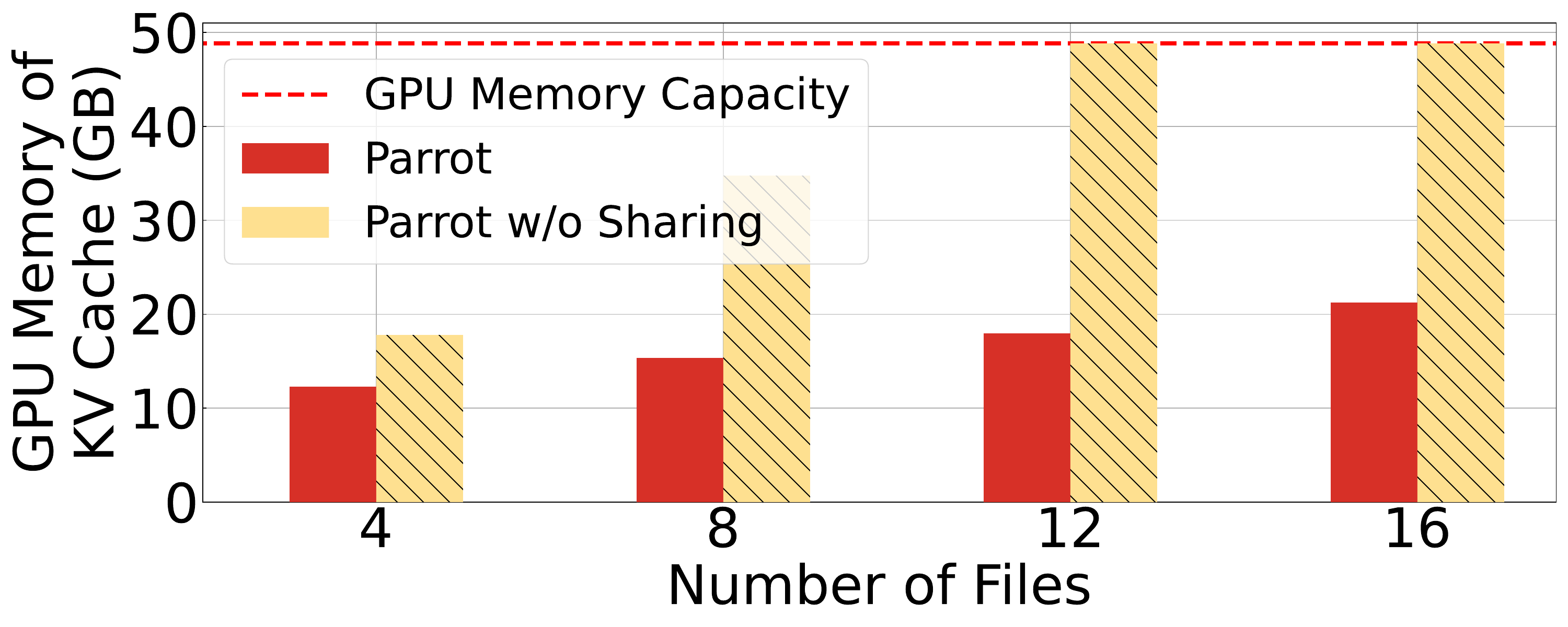}\\\vspace{-1mm}
        \caption{GPU Memory of KV Cache\label{fig:expt_multi_mem}}
    \end{subfigure}
    \caption{The latency and memory usage for multi-agent programming, with varying number of files to program. \label{fig:expt_multi_agent}}
\end{figure}

We assess the performance of multi-agent systems utilizing MetaGPT~\cite{metagpt} within \sysname{}. A workflow is constructed with three distinct roles. Initially, the Architect outlines the project's file structures and specifies APIs within each file for a given task. Subsequently, multiple Coders undertake the project implementation, with each focusing on a specific file. Following the integration of the code from all files, several Reviewers engage in the process, each examining and commenting on a single file. The Coders then revise their code based on these comments. This review-and-revision cycle is iterated three times to produce the final code. \autoref{fig:expt_multi_agent} illustrates the latency and memory consumption of \sysname{} compared to baseline systems on one A100 running LLaMA 13B. \sysname{} achieves a speedup of up to $11.7\times$ compared with the latency-centric baseline. The primary improvement is attributed to \sysname{}'s capability to deduct the performance objectives for LLM requests based on the end-to-end performance criteria. For this specific multi-agent scenario, the goal is to minimize the time taken to deliver the final code. \sysname{} identifies multiple task groups within the parallel processes of coding, reviewing, and revising, facilitating larger batch sizes to enhance throughput and reduce the completion time of task groups. We also contrast \sysname{} with an throughput-centric baseline that uses larger batch on purpose to optimize cluster throughput, which also shows higher concurrency and better completion time than the latency-centric baseline.
  
Even when compared to the throughput-centric baseline, \sysname{} demonstrates superiority, being faster by up to $2.45\times$. This enhancement \cfnew{mainly} stems from \sysname{}'s ability to decrease redundancy through its prompt structure analysis\cfnew{, which contributes a $2.35\times$ acceleration}. Given the interactive nature of the roles in MetaGPT, there is considerable overlap in the context among different roles, which \sysname{} capitalizes on by sharing this common context as a prompt prefix. The static prefix sharing mechanism from vLLM does not work in this dynamic scenario. Without a grasp of the prompt's structure, it cannot identify dynamically generated \sv{}s that could also be shared during runtime. 
As depicted in \autoref{fig:expt_multi_mem}, \sysname{} without this sharing capability would hit the GPU memory ceiling. Additionally, \sysname{}'s specialized GPU kernel for processing the shared prefix achieves a further $1.2\times$ speedup when there are 16 files, compared to using vLLM's PagedAttention, due to the reduced memory transactions.  

\subsection{Scheduling of Mixed Workloads}
To assess the performance of \sysname{} on a multi-GPU setup, we configure a cluster with four A6000 (48GB) GPUs, each hosting a separate LLM engine (LLaMA 7B), resulting in a total of four LLM engines. We emulate a real-world scenario where LLM services encounter a variety of demands by injecting a mix of requests from chat applications at a rate of 1 req/s and from data analytic tasks (i.e., map-reduce applications) previously analyzed in \S\ref{sec:eval_summary}. Requests from the chat applications are characterized by their need for low latency, whereas the map-reduce applications prioritize high throughput, creating a challenge when they are concurrently processed by the same LLM engine. We benchmark \sysname{} against two reference implementations: one tailored for latency, limiting engine capacity to reduce decoding time, and another for throughput, utilizing full engine capacity to maximize GPU utilization.  
  
\begin{figure}[t]
    \centering
    \includegraphics[width=1\linewidth]{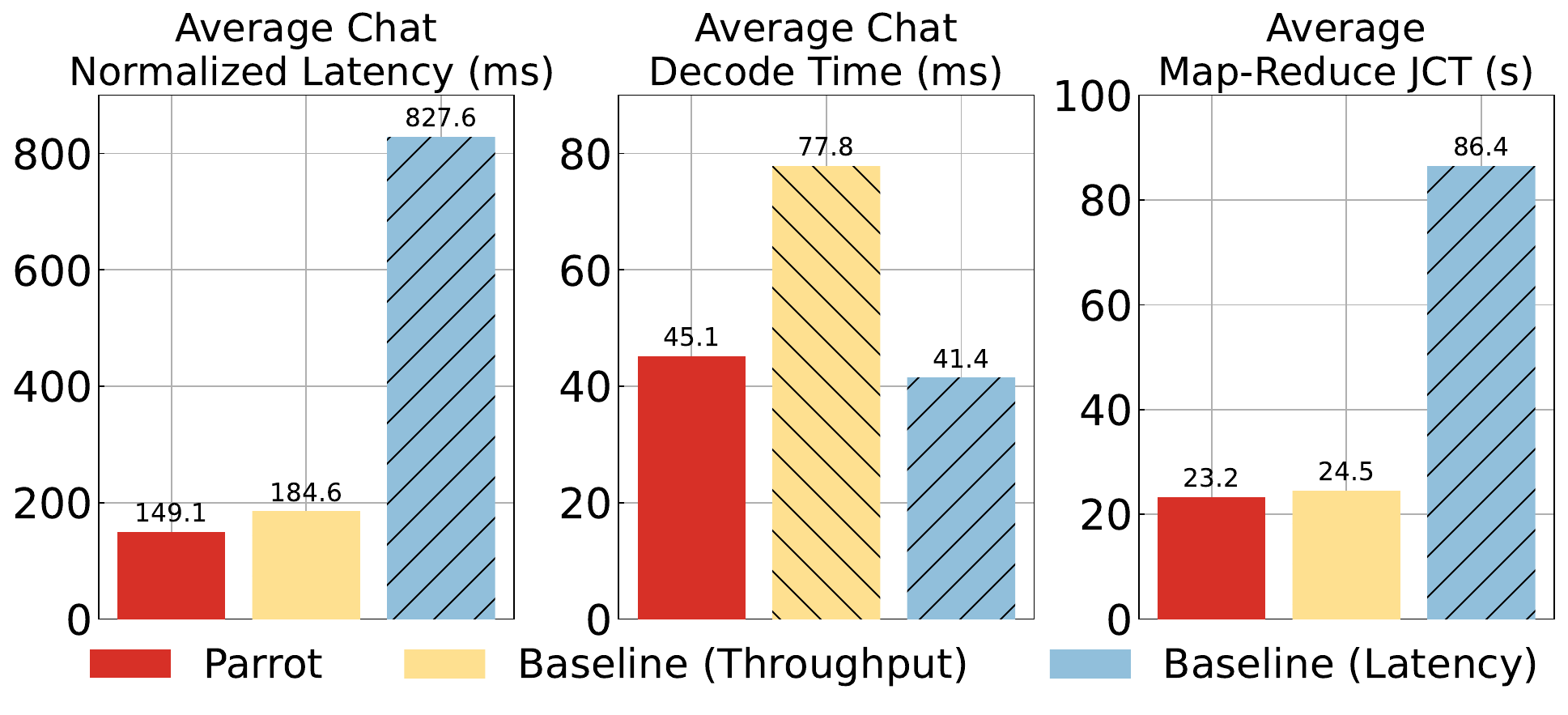}
    \caption{The mixture of chat and map-reduce applications.}
    \label{fig:expt_mixed}
\end{figure}

The results depicted in \autoref{fig:expt_mixed} demonstrate that \sysname{} attains a $5.5\times$ and $1.23\times$ improvement in normalized latency (measured as request latency per number of output tokens)~\cite{orca,vllm} for chat applications in comparison to the latency-focused and throughput-focused baselines, respectively. In terms of token generation speed for chat applications, \sysname{} delivers performance on par with the latency-centric baseline and outperforms the throughput-centric baseline by $1.72\times$. For map-reduce applications, \sysname{} reaches a $3.7\times$ speedup over the latency-centric baseline and is $1.05\times$ more efficient than the throughput-centric baseline. \sysname{} excels by providing both low latency for chat applications and high throughput for map-reduce applications. It mitigates the contention between chat and map-reduce workloads by intelligently scheduling them on separate engines. These findings underscore the significance of specialized handling for diverse requests to enhance the overall performance of LLM services.  

\section{Related Works}

\paragraph{Deep Learning Serving Systems.} The field of model serving has seen a surge of research activity in recent years, with many systems developed to address the different challenges of deep learning model deployment. The systems include Clipper~\cite{clipper}, TensorFlow Serving~\cite{tfserving}, Clockwork~\cite{clockwork}, REEF~\cite{reef}, AlpaServe~\cite{alpaserve},  which have explored many aspects including batching, caching, placement, scheduling, model parallelism for the serving of single or multiple models. These systems were proposed for serving general deep learning models, which have less consideration about the unique requirements of large language models, e.g., autoregressive decoding. Orca~\cite{orca} proposed a fine-grained scheduling mechanism that can batch multiple LLM requests at the iteration level, which is also known as continuous batching. vLLM proposes PagedAttention~\cite{vllm} allows the batching of LLM requests with different lengths using non-contiguous memory, increasing memory utilization. These systems for LLM serving still treat LLM requests separately, missing the opportunities to understand the interconnections within an application and exploit the commonality of different requests. \sysname{} is orthogonal to them. With more application-level knowledge exposed by \sv{}s, \sysname{} can do data flow analysis on LLM requests, which enables a brand new optimization space with the final goal of optimizing the end-to-end performance of applications, rather than individual requests. 

\paragraph{LLM Orchestrator Frameworks.} LLM orchestration frameworks help developers create and manage applications powered by LLMs. They simplify the process of prompt design, and orchestration of multiple LLM requests, which enable developers to interact with LLMs easily. LangChain~\cite{langchain} is a Python framework that provides many workflow patterns, e.g., chain, map-reduce so that developers can easily customize their own LLM applications. Semantic Kernel~\cite{sk} introduces Planners are semantic agents that can automatically generate plans based on the needs of the users. PromptFlow~\cite{promptflow} supports chains of native and semantic functions and visualizes them as a graph. LlamaIndex~\cite{llamaindex} allows developers to use natural language queries to retrieve relevant documents. 
\sysname{} is orthogonal to these frameworks and can be easily integrated with these frameworks to support \sysname{}’s APIs with \sv{} abstraction, as discussed in \S\ref{sec:discussion}.


\paragraph{DAG-aware System Optimizations.} Dependency graphs or DAGs (Directed Acyclic Graphs) widely exist in many kinds of systems, and many optimizations have been proposed to optimize the systems by exploiting the DAG information. Tez~\cite{tez}, Dryad~\cite{dryad}, and Graphene~\cite{graphene} use the task dependency to optimize the scheduling and packing of parallel data analytic workloads. SONIC~\cite{sonic}, Caerus~\cite{caerus}, and Orion~\cite{orion} optimize serverless functions from the aspects of communication, latency, and cost. \sysname{} learns from the previous system works and realizes the importance of correlations of LLM requests to optimize the end-to-end performance of LLM applications. This motivates \sysname{} to build APIs for exposing such dependency information. Moreover, it is unique to LLM applications to understand the prompt structure in addition to request-level dependency, which is necessary for communication and identifying commonality across LLM requests. This motivates us to propose the \sv{} abstraction, instead of just using a DAG of requests.
\section{Conclusion}
This paper proposes \sysname{} that treats LLM applications as first-class citizens and targets to optimize the end-to-end performance of LLM applications, instead of only optimizing individual LLM requests. We propose \sv{} as the key abstraction that exposes the dependency and commonality of LLM requests, enabling a new optimization space. Our evaluation shows \sysname{} can optimize LLM-based applications by up to $11.7\times$. We envision this new angle of efficiency improvement of LLM applications 
brings a broad future direction to study other scheduling features like the fairness of \emph{end-to-end} performance of LLM applications. 


\section*{Acknowledgments}
We thank the anonymous reviewers and the shepherd for their constructive feedback and suggestions. Zhenhua Han, Yuqing Yang and Chen Chen are the corresponding authors.

\bibliographystyle{plain}
\bibliography{ref}

\end{document}